\documentclass[%
 aip,
 amsmath,
 amssymb,
 reprint,
]{revtex4-1}

\usepackage{graphicx}
\usepackage{dcolumn}
\usepackage{bm}

\usepackage[utf8]{inputenc}
\usepackage[T1]{fontenc}
\usepackage{mathptmx}
\usepackage{etoolbox}
\usepackage{enumitem}
\usepackage{float}

\usepackage{hyperref}
\usepackage[ruled,vlined]{algorithm2e}
\SetKwInput{KwInput}{Input}                
\SetKwInput{KwOutput}{Output}              
\usepackage[usenames,dvipsnames]{xcolor}

\newcommand{\R}{\mathbb{R}}

\newcommand{\vthl}{v_\textrm{thl}}
\newcommand{\vthh}{v_\textrm{thh}}

\begin{document}

\title[Differentiating Ring Oscillator Lattices]{Transient Dynamics in Lattices of Differentiating Ring Oscillators}

%
\author{Peter DelMastro}
\email[]{pdelmastro@vt.edu}
\affiliation{Department of Mathematics, Virginia Tech}
\author{Arjun Karuvally}
\email[]{akaruvally@cs.umass.edu}
\affiliation{Manning College of Information \& Computer Sciences, University of Massachusetts Amherst}
\author{Hananel Hazan}
\email[]{hananel@hazan.org.il}
\affiliation{Allen Discovery Center, Tufts University}
\author{Hava Siegelmann}
\email[]{hava@cs.umass.edu}
\affiliation{Manning College of Information \& Computer Sciences, University of Massachusetts Amherst}
\author{Edward Rietman}
\email[]{erietman@gmail.com}
\affiliation{Manning College of Information \& Computer Sciences, University of Massachusetts Amherst}

\date{\today}

%
%
\begin{abstract}
Recurrent neural networks (RNNs) are machine learning models widely used for learning temporal relationships.
Current state-of-the-art RNNs use integrating or spiking neurons --- two classes of computing units whose outputs depend directly on their internal states --- and accordingly there is a wealth of literature characterizing the behavior of large networks built from these neurons.
On the other hand, past research on \textit{differentiating neurons}, whose outputs are computed from the derivatives of their internal states, remains limited to small hand-designed networks with fewer than one-hundred neurons.
Here we show via numerical simulation that large lattices of differentiating neuron rings exhibit local neural synchronization behavior found in the Kuramoto model of interacting oscillators.
We begin by characterizing the periodic orbits of uncoupled rings, herein called \textit{ring oscillators}.
We then show the emergence of local correlations between oscillators that grow over time when these rings are coupled together into lattices.
As the correlation length grows, transient dynamics arise in which large regions of the lattice settle to the same periodic orbit, and thin domain boundaries separate adjacent, out-of-phase regions. 
The steady-state scale of these correlated regions depends on how the neurons are shared between adjacent rings, which suggests that lattices of differentiating ring oscillator might be tuned to be used as reservoir computers.
Coupled with their simple circuit design and potential for low-power consumption, differentiating neural nets therefore represent a promising substrate for neuromorphic computing that will enable low-power AI applications.

\end{abstract}
\maketitle

%
%
\begin{quotation}
Recurrent neural networks are widely used machine learning models for learning temporal relationships. Like other modern neural networks, they are power hungry, so there is a rising interest for low-power alternatives.
A primary source of this high power utilization is the use of integrating neurons as the fundamental computational units, as these neurons require steady currents to maintain their internal states.
There exists, however, a largely forgotten class of neurons, known as \textit{differentiating neurons}, which exhibit oscillatory, more energy efficient dynamics when they are organized into rings. 
Small networks of these \textit{ring oscillators} have only been used in small-scale applications like locomotion in robots.
Here, we extend past research and characterize the behavior of coupled ring oscillator lattices with thousands of rings from the perspective of oscillator synchronization. 
We find in simulation that, similar to the Kuramoto model, these lattices develop locally synchronized regions whose sizes depend on how the rings are coupled together. This property suggests that lattices of differentiating ring oscillators might be tuned to be used as low-power reservoir computers.
\end{quotation}

%
%
\section{Introduction}

Artificial neural networks form a diverse family of biologically inspired machine learning techniques, with recurrent architectures \citep{elman90, hochreiter97, Erichson2021LipschitzRN, Rusch2020CoupledOR, Rusch2022LongEM, Gu2022S4} being particularly well-suited for modeling temporal relationships. Their ability to store and manipulate external inputs over time have made them popular in sequence processing tasks such as time series prediction \citep{Chandra2021EvaluationOD, Pathak2018ModelFreePO}, language modeling \citep{Melis2017OnTS, Merity2017RegularizingAO, Gu2023MambaLS}, and control \citep{Meng2021MemorybasedDR}.

\begin{figure}[t!]
    \includegraphics[width=0.95\columnwidth]{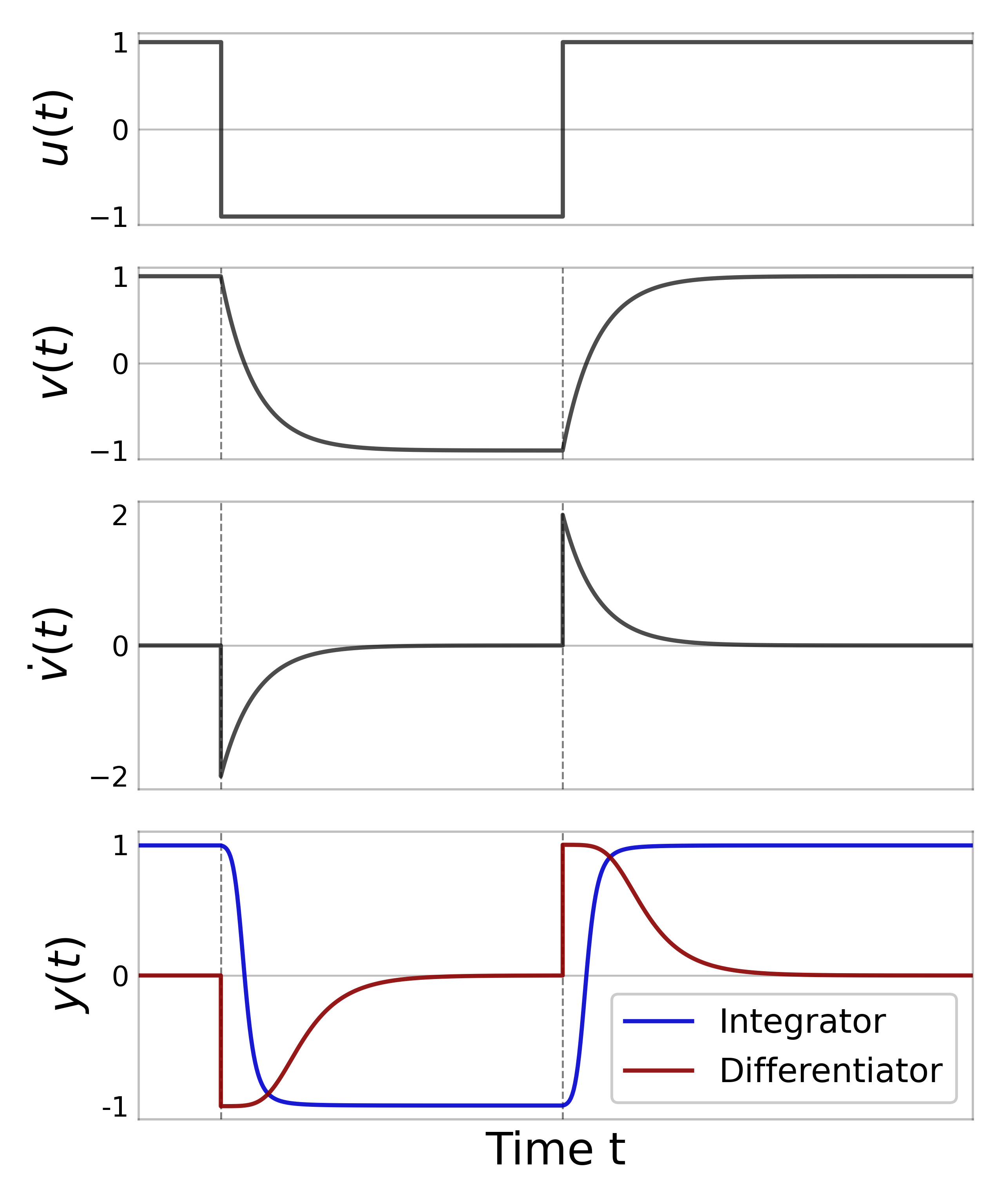}
    \caption{
    \label{fig:int-vs-diff-neurons}
    \textbf{Integrating vs. Differentiating Neurons} \ \ 
	This figure shows the outputs $y(t)$ of an integrating neuron and a differentiating neuron in response to an example binary (+1/-1) input signal $u(t)$. The internal state $v(t)$ of both neurons is the same and evolves in time to minimize the difference $u(t) - v(t)$. The output of the integrator, shown blue on the bottom plot, depends on $v(t)$ as $y_\textrm{int}(t) = \tanh( \beta v(t))$ with $\beta = 3$, so this neuron's output resembles a slightly time-shifted copy of the input signal. The differentiator's output, shown in red in the same plot, is computed from \textit{time derivative} of $v(t)$; in particular, $y_\textrm{diff}(t) = \tanh(\beta \dot{v}(t))$. The neuron's output is large in magnitude when $\dot{v} = (u - v)/\tau$ is large in magnitude, like when $u(t)$ switches sign, so the differentiator responds to rapid changes in input.
}
\end{figure}

Despite the diversity of modern neural architectures available for applications, RNNs are built with \textit{integrating} neurons --- computing units whose outputs are increasing functions of their internal states that require steady currents to maintain. A classic example is the \textit{leaky integrator} \citep{Jaeger2007LeakyIntegrator} described by the differential equations
\begin{equation}
    \tau \dot{v}(t) = -v(t) + u(t), \quad \  y(t) = \phi \left(v(t)\right)
\end{equation}
where $v$ denotes the internal (hidden) state of the neuron, $u$ the aggregate input to the neuron, and $y$ the output of the neuron. The time-evolution of the internal state $v(t)$ can be written as a convolution
\begin{equation}
v(t) = e^{-t/\tau}v(0) + \int_{0}^t u(t-s) e^{-(t-s)/\tau} ds
\end{equation}
so $v$ is an exponentially \textit{decaying} history of the input sequence. Standard activation functions are typically monotone increasing, so integrating neurons produce a large output in response to sustained high input. See Fig. \ref{fig:int-vs-diff-neurons} for an example of this behavior for the activation function $\phi(x) = \tanh(\beta x)$.

The decaying internal memory imbues the neuron with the ability to forget past inputs, and when this is coupled with nonlinear self feedback, such neurons can store information over long periods of time \cite{Hopfield1982NeuralNA, Krotov2021LargeAM, Karuvally2022EnergybasedGS}. Such long-term memory is a blessing computationally but it can also be curse, for instance due to vanishing and exploding gradient problems during training \citep{Doya1993, bengio94, Pascanu2012OnTD}.
This internal memory also requires steady currents to maintain, making these neurons require large amounts of power during computation \cite{Stiefel2023TheEC}.

\textit{Differentiating neurons} \citep{Hasslacher97,Rietman03}, herein referred to as \textit{differentiators}, are a natural counterpart to integrators. As their name suggests, the output of a differentiator is computed from the \textit{derivative} of its internal state, i.e.
\begin{equation}
    y = \phi(\tau \dot{v}) = \phi(-v + u)
\end{equation}
These neurons therefore respond to \textit{changes in input}, such as in Fig. \ref{fig:int-vs-diff-neurons}. This property causes networks of differentiating neurons to exhibit ``event-based'' dynamics, wherein short pulses of neural activity travel around the network from neuron to neuron \citep{Rietman03}. Such dynamics are energy efficient as the neurons spend most of their time dormant and become active only for short periods of time in response to input changes. Coupled with their simple circuit design, differentiating neural nets are a promising substrate for neuromorphic computing.

Networks of differentiating neurons have existed for decades and were at one point competitive with feedforward integrating networks. When used in tandem with integrating neural nets, these networks provided novel, minimal complexity approaches to satellite control \citep{SATBOT1995}, basic motion tracking \citep{Frigo98}, and central pattern generation in small robots \citep{Hasslacher95, Still98}. After the introduction of the backpropagation algorithm, the popularity of differentiating neuron networks began to decline, and since then, they have existed in niche applications. For instance, when used in conjunction with integrating neurons, very small networks ($<$ 50 neurons) have achieved behavior such as voice and feature recognition. However, applications of differentiating neurons have to our knowledge always been designed by hand, and this limits the size of the networks in practice. 

To enable the investigation of much larger networks, we introduce an event-based simulation for these hardware neural networks. We then use this simulation engine to numerically characterize the dynamics of large differentiating neural networks on the order of tens of thousands of neurons. 

We begin by studying the dynamics of rings of neurons, referred to as \textit{ring oscillators} \citep{Rietman03}. These cyclic networks were previously shown in hardware to naturally exhibit oscillations \citep{Rietman03}. We find that, in simulation, the number of stable periodic orbits of these ring oscillators grows linearly in their ring size (cf. Section \ref{sec:ring-oscillator-periodic-orbits}). Although rings of integrating neurons also generate complex behavior \citep{Rodan2011MinimumComplexESN}, we find that the stable periodic orbits of differentiating ring oscillators have lower complexity, with the output of each neuron being identical to all other neurons in the ring modulo phase shifts. Each periodic orbit is characterized by a different number of pulses --- short bursts of non-zero neuronal activity --- traveling around the ring. 

Our subsequent simulations reveal that more complex dynamics can be achieved by coupling ring oscillators into two-dimensional rectangular lattices by sharing neurons between rings (cf. Section \ref{sec:lattices}). This complexity is not in the number of stable periodic orbits; rather, complexity emerges at a macroscopic level in both transient and steady-state dynamics. In homogeneous lattices where the coupling between rings is uniform across the network, we find that their dynamics resembles magnetic domains \citep{Sethna2021StatMechCorrFunc}: the rings converge towards a coupling-dependent periodic orbit, and the network organizes into local domains that are out of phase with each other.
As these domains try to synchronize, (quasi)periodic structures emerge at their boundaries. 

We employ oscillator phase reduction analysis \citep{Nakao2016PhaseReduction} to extract correlation lengths \citep{Sethna2021StatMechCorrFunc} characterizing the scale of these locally synchronized regions. 
This analysis reveals that both the scale and structure of these phase-correlated regions in the steady-state, as well as the rate at which they develop, vary widely with the coupling geometry, e.g. the way neurons are shared between rings. 
Indeed, for certain geometries, rings of small and large sizes are capable of achieving nearly global oscillator synchronization, whereas for other couplings, these correlations saturate and global synchronization is never reached. 

The rest of this paper is organized as follows: In Section \ref{sec:background}, we review past literature on rings of differentiating neurons, oscillator phase reduction for dynamics nearby periodic orbits, and phase synchronization in coupled oscillator networks. We then introduce in Section \ref{sec:nv-nets-formalism} our event-based formalism for simulating networks of differentiating neurons, and we use this formalism to characterize the periodic orbits of individual differentiating ring oscillators in Section \ref{sec:ring-oscillator-periodic-orbits}. Finally, in Section \ref{sec:lattices}, we present numerical simulation of coupled differentiating ring oscillator lattices, using the technique of oscillator phase correlation to characterize the scale of oscillator synchronization as a function of ring connectivity.

%
%
\section{Background}
\label{sec:background}

\subsection{Differentiating Neural Ring Oscillators}
\label{sec:background-ring-oscillators}

Past studies of differentiating neurons have placed emphasis on \textit{rings} of neurons \cite{Rietman03, Hasslacher95, Hasslacher97}, the minimal architecture required for sustained dynamical activity. These studies observed that rings can settle to a number of periodic orbits that function as a form of memory.

An $n$-ring oscillator has $n$ neurons organized as a ring, with the output of the $i$-th neuron acting as the input to the $(i+1)$-st neuron (modulo $n$). The system is therefore described by the differential equations
\begin{equation}
\tau_i \dot{v}_i = y_{i-1}(t) - v_i(t) \quad i = 1, 2, ..., n
\label{eqn:ring-oscillator-ode}
\end{equation}
where $y_i(t) = \phi(\tau_i \dot{v}_i)$ is the output of neuron $i$ and we consider $i-1$ modulo $n$. When the activation function $\phi$ is an inverting Schmitt trigger, these neurons can transmit pulses to their children, so with neurons organized into rings, these pulses may never die out and move indefinitely from neuron to neuron around the ring \cite{Rietman03}.

The state of an $n$-ring oscillator is characterized by the capacitor voltages $\mathbf{v} = (v_1, ..., v_n)$ with each $v_i \in [0,1]$ and output voltages $\mathbf{y} = (y_1, ..., y_n)$ with each $y_i$ taking binary value $0$ or $1$. 
To better model the hardware (c.f. Section \ref{sec:nv-nets-formalism}), the output $\mathbf{y}$ is restricted to the set of binary vectors with no adjacent 1s, where the first and last indices are considered to be adjacent. For instance, the valid output states for a 4-ring oscillator are $(0,0,0,0), (0,0,0,1), (1,0,1,0)$, and all cyclic permutations of these vectors. In general, the number of equivalence classes under rotation of these valid output states, denoted here by $V_n$, is given by the number of binary necklaces with no 11 subsequence:\citep{Rietman03}
\begin{equation}
V_n = \frac{1}{n} \sum_{d|n} \varphi(d) \big[F(d-1) + F(d+1)\big]
\end{equation}
Here, $\varphi$ is the Euler-phi function and $F(n)$ is the $n^{th}$ Fibonacci number. The number of equivalence classes grows exponentially with $n$. 

Empirical studies \cite{Rietman03} of differentiating ring oscillators in hardware reported that these networks support a \textit{periodic orbit} for each equivalence class of valid strings under rotation. 
By periodic orbit, we are referring to a trajectory that satisfies the differentiation equations (\ref{eqn:ring-oscillator-ode}) and is periodic in time, say $\mathbf{v}(t) = \mathbf{v}(t+P)$ for all $t$ some oscillation period $P$. 
The periodic orbits of the ring oscillator were previously reported to be stable solutions, but we find in Section \ref{sec:ring-oscillator-periodic-orbits} that only $\lfloor n/2 \rfloor$ of these orbits are stable attractors.
Still, periodic orbits are undoubtedly the steady-state behavior of rings of differentiating neurons. This property justifies the terminology \textit{ring oscillator} for these circuits.

\subsection{Oscillator Phase Reduction}
\label{sec:background-phase-reduction}

Phase reduction \citep{Nakao2016PhaseReduction} is a technique to describe the dynamics of a system nearby a stable periodic orbit by reducing oscillations in multiple dimensions to a single periodic variable $\theta$ called the \textit{phase}. The phase changes with constant rate in time, for instance $\theta(\mathbf{x}(t)) = \theta(\mathbf{x}_0) + t / P \bmod{1}$ where $P$ is the period of the orbit.
This approach to dimensionality reduction is particularly useful in characterizing the complex dynamics of coupled oscillator networks, as discussed in Sec. \ref{sec:background-oscillator-networks}. 

To perform phase reduction, we first assigns a phase $\theta$ to each point $\mathbf{x}$ in the state space of the system. We first define a mapping between the periodic orbit $\Gamma$ of interest and the interval $[0,1)$. Fixing a reference state $\mathbf{x}_0 \in \Gamma$ along the orbit, we define the phase of this point as $\theta(\mathbf{x}_0) = 0$. The phase of other states on $\Gamma$ are defined using the flow map $\Phi: \R^n \times \R \to \R^n$ induced by the dynamical system; $\Phi(\mathbf{x},t)$ denotes the state of the system at time $t > 0$ if the system started in state $\mathbf{x}$ at time $t = 0$. Each state $\mathbf{x} \in \Gamma$ can be uniquely written as $\mathbf{x} = \Phi(\mathbf{x}_0, t)$ for some $t \in [0, P)$, and to this state, we assign the phase
\begin{equation}
    \theta(\mathbf{x}) = t / P \ , \quad \textrm{for} \ \mathbf{x} = \Phi(\mathbf{x}_0, t) \in \Gamma
\end{equation}
For points $\mathbf{x} \notin \Gamma$ but within the basin of attraction of the period orbit, we can then define
\begin{equation}
\theta(\mathbf{x}) = (\theta \circ \gamma)(\mathbf{x}) \quad \textrm{where} \quad \gamma(\mathbf{x}) = \lim_{n\to \infty} \Phi(\mathbf{x}, n P)
\label{eqn:phase-reduction}
\end{equation}
This definition assigns to each state $\mathbf{x}$ the phase of the point $\gamma(\mathbf{x})$ on the periodic orbit that we return to whenever $t \equiv 0 \bmod{P}$ in the limit of $t \to \infty$. This choice leads to the desired property $\theta(\mathbf{x}(t)) = \theta(\mathbf{x}(0)) +  t / P \bmod{1}$ due to the semi-group property of the flow: $\Phi\big(\Phi(\mathbf{x},t'),t\big) = \Phi\big(\Phi(\mathbf{x},t),t'\big)$, so that
\begin{align*}
\gamma\big(\Phi(\mathbf{x},t)\big) &= \lim_{n \to \infty} \Phi\big(\Phi(\mathbf{x},t), n P\big) = \lim_{n \to \infty} \Phi\big(\Phi(\mathbf{x},nP), t\big) \\
&= \Phi\big(\gamma(\mathbf{x}), t\big) 
\end{align*}
assuming continuity of $\Phi$ in $\mathbf{x}$. This implies
\begin{equation}
\theta\big(\Phi(\mathbf{x},t)\big) = \theta\big(\Phi(\gamma(\mathbf{x}),t)\big) = \theta \big(\gamma(\mathbf{x})\big) + t/P \bmod{1}
\end{equation}
The \textit{difference} in phase between two trajectories therefore invariant in time: starting from initial states $\mathbf{x}$ and $\mathbf{x'}$,
\begin{equation}
\theta\big(\Phi(\mathbf{x},t)\big) - \theta\big(\Phi(\mathbf{x}',t)\big) \equiv \theta(\mathbf{x}) - \theta(\mathbf{x}') \bmod{1}
\end{equation}

\begin{figure}[t!]
    \includegraphics[width=0.9\columnwidth]{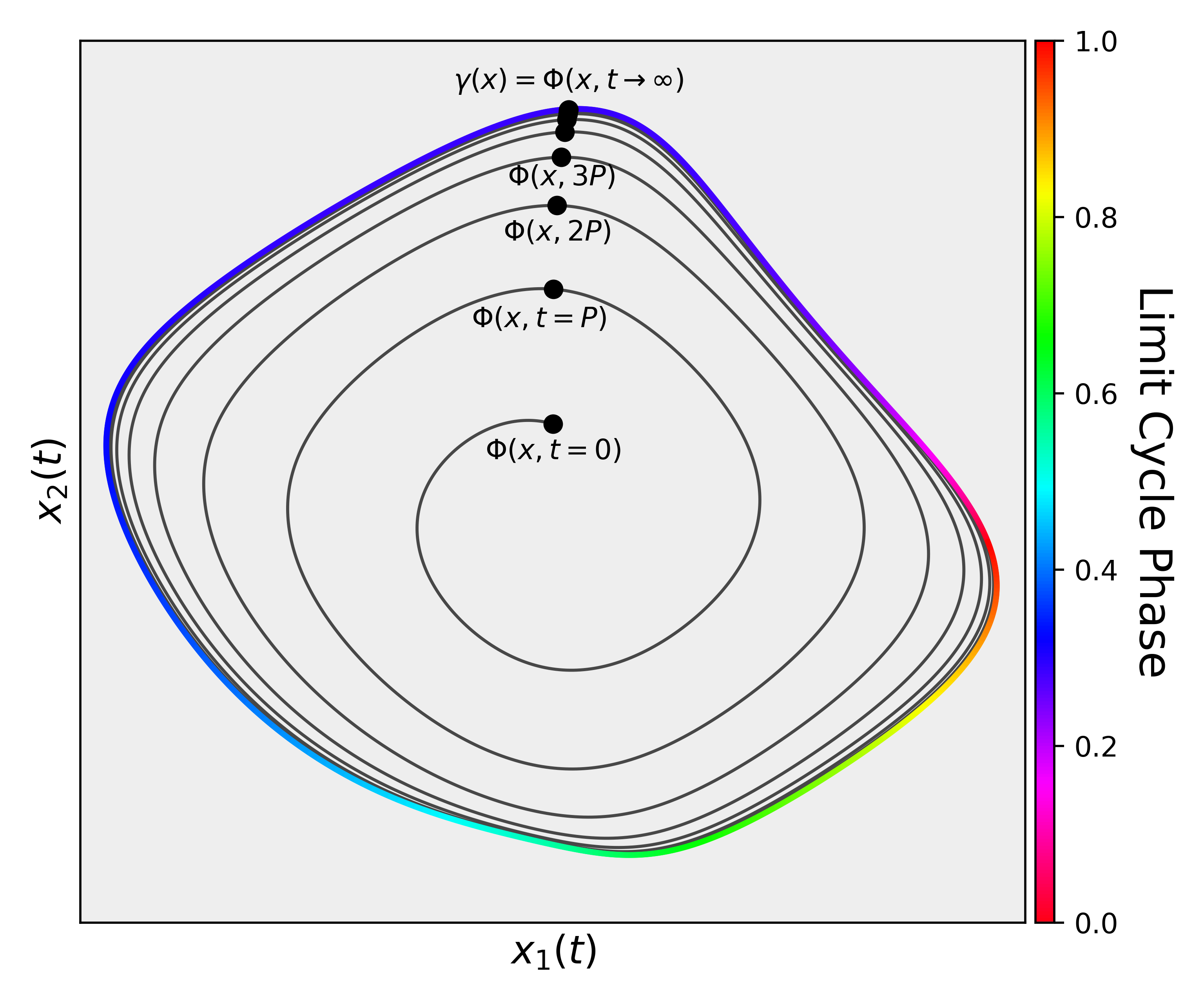}
    \caption{
    \label{fig:phase-reduction}
    \textbf{Oscillator phase reduction} of a hypothetical limit cycle oscillator. The system here two variables $(x_1,x_2)$ and we've given it a periodic orbit its phase space, indicated by the colored closed curve $\Gamma$. Each point in the loop is assigned a unique phase value $\theta \in [0,1)$ as indicated by the cyclical colormap.
    Notice that the oscillator phase, which changes at a constant rate in time, does not change at a constant rate with respective to space --- spatially, a majority of the cycle corresponds to phase between 0.2 and 0.4, indicating a region where the trajectory has high velocity $\mathbf{x}'(t)$.
    The black curve indicates a trajectory of the system starting from a state $\mathbf{x} \in \R^2$. Along the trajectory we mark the points $\Phi(\mathbf{x}, nP)$ to indicate that they converge towards a point on the periodic orbit $\Gamma$. The phase $\theta(\mathbf{x})$ assigned to $\mathbf{x}$ is the phase of the point $\gamma(\mathbf{x}) \in \Gamma$ to which the trajectory starting from $\mathbf{x}$ converges.
}
\end{figure}

Fig. \ref{fig:phase-reduction} shows an example of this definition for a hypothetical two-dimension system. The periodic orbit $\Gamma$ is denoted by the colored curve, with the color indicating the phase $\theta$ of each point along $\Gamma$. A trajectory starting from an arbitrary $\mathbf{x}$ is also shown, with the points $\Phi(\mathbf{x},nP)$ converging to a point $\gamma(\mathbf{x})$ on the periodic orbit. The phase of $\gamma(\mathbf{x})$ is therefore the phase assigned to $\mathbf{x}$, i.e. $\theta(\mathbf{x}) = \theta(\gamma(\mathbf{x}))$.

Analytic phase reduction is particularly challenging in the case of differentiating neuron rings, especially because these oscillators support \textit{multiple} periodic orbits, which we characterize in Section \ref{sec:ring-oscillator-periodic-orbits}.
In this work, we therefore take a numerical approach to phase reduction, computing oscillator phase directly rather than deriving an analytic expression for the phase dynamics.
We explain our procedure for numerical oscillator phase analysis in Section \ref{sec:lattices-measuring-correlations} when we employ this technique to characterize synchronization in networks of coupled ring oscillators.

\subsection{Synchronization in Coupled Oscillator Networks}
\label{sec:background-oscillator-networks}

Phase reduction can be particularly informative in characterizing the behavior of \textit{networks} of oscillators. 
Consider a collection of independent systems $\mathbf{x}_1, ..., \mathbf{x}_s$ described by the differential equations $\dot{\mathbf{x}}_i = \mathbf{F}_i(\mathbf{x}_i)$, each of which supports its own period orbit. 
We might then modify the dynamics of these systems by introducing interactions between them, such as  
\begin{equation}
    \dot{\mathbf{x}}_i = \mathbf{F}_i(\mathbf{x}_i) + \epsilon \mathbf{C}_i(\mathbf{x}_1, ..., \mathbf{x}_s)
\end{equation}
When $\epsilon > 0$, the added interactions perturb the periodic orbits of the individual systems, giving coupled oscillator networks the potential to exhibit wide ranges of dynamical features.

\textit{Synchronization} is a particularly well-studied feature of coupled oscillator networks. Letting $\theta_i$ denote the phase reduction map for the $i$-th system, we say oscillators $i$ and $j$ in the network are phase synchronized if their phases align in time: $\theta_i(\mathbf{x}_i(t)) = \theta_j(\mathbf{x}_j(t))$ for all $t$ in some interval.  
Phase reduction can provide analytic solutions for the dynamics for some classes of coupled oscillator networks, for example, the Kuramoto model. Under certain assumptions on $\mathbf{F}_i$ and $\mathbf{C}_i$, the \textit{Kuramoto model} \citep{Kuramoto1975Original, Kuramoto1984ChemicalOW} \begin{equation}
    \dot{\theta}_i = \omega_i + \sum_{j=1}^s k_{ij} \sin(\theta_j - \theta_i)
\end{equation}
well approximates the dynamics of the phase variables $\theta_i(t) := \theta_i(\mathbf{x}_i(t))$. 
Here, $\omega_i$ denotes the oscillation frequency $\omega_i = 2\pi/P_i$ of the $i$-th oscillator, and $k_{ij}$ is the coupling strength between the $i$-th and $j$-th oscillators. The latter coefficients determine the topology of the oscillator network; for instance $k_{ij} = k \in \mathbb{R}$ for all $i,j$ denotes a fully and uniformly connected oscillator network.

The Kuramoto model has been studied on many network structures (cf. reviews \citep{Acebrn2005KuramotoRev1, Rodrigues2015KuramotoRev2}). 
Two-dimensional rectangular lattices \citep{Sarkar2021KuramotoLattice} are of particular interest in our study. For such networks one has a grid oscillators $\theta_{i,j}$, $i,j = 1, ..., n$ with dynamics determined by the equations
\begin{align}
    \dot{\theta}_{ij} = 
    \omega + k \big[ &\sin(\theta_{i,j+1} - \theta_{ij}) + \sin(\theta_{i,j-1} - \theta_{ij}) \\ 
    + & \sin(\theta_{i+1,j} - \theta_{ij}) + \sin(\theta_{i-1,j} - \theta_{ij})  \big] \nonumber
\end{align}
Above, the oscillator at lattice site $(i,j)$ interacts with only the oscillators at the four neighboring lattice sites. We've also used a uniform oscillation frequency $\omega$ and coupling strength $k$. Under change of variables to a suitable rotating frame of reference, one can show it suffices to take $\omega = 0$ and $k = 1$, which becomes equivalent to the 2D XY model when a stochastic driving force is added to the system \citep{Sarkar2021KuramotoLatticeWithNoise}.

Such uniform Kuramoto oscillator lattices exhibit two types of steady-state behavior \citep{Sarkar2021KuramotoLattice}: 
\begin{itemize}
    \item \textit{Global synchronization}, wherein all oscillators have the same phase: for all $i,j,k,l$, 
    \begin{equation*}
        \theta_{ij}(t) = \theta_{kl}(t)
    \end{equation*}
    \item \textit{Phase-locking}, where the \textit{difference in phase} between any pair of oscillators is fixed in time: for all $i,j,k,l$,  
    \begin{equation*}
        \frac{d}{dt} |\theta_{ij}(t) - \theta_{kl}(t) | = 0
    \end{equation*}
\end{itemize}
Starting from non-equilibrium states, these oscillator lattices \textit{relax} to one of these steady states. The lattice dynamics leading to this convergence, called the \textit{relaxation} or \textit{transient} dynamics, involves the formation of small clusters of locally synchronized oscillators. The scale of these clusters grows in time, ultimately leading to a single globally synchronized cluster or non-zero gradients in oscillator phases that stabilize to a phase-locked state. 

These transient dynamics can be characterized in terms of vortices and anti-vortices moving through the lattice \citep{Sarkar2021KuramotoLattice}. These so called \textit{defects} can collide with each other and annihilate, resulting in the formation of a larger region of synchronized oscillators. The globally synchronized state occurs if all defects annihilate, whereas some of these structures still persist in phase-locked states.
We will show that we show below that for differentiating ring oscillator networks are similar to Kuramoto lattices in this sense, capable of forming globally synchronized states and out-of-phase domains depending on the lattice coupling.

%
%
\section{Event-Based Simulation for Networks of Differentiating Neurons}
\label{sec:nv-nets-formalism}

In the past, ring oscillators have been coupled together into networks through additional resistors \citep{Rietman03}. There are certainly other choices to perform this coupling; for instance, our work builds lattice networks by placing a ring oscillator at each lattice site and coupling rings by \textit{sharing} neurons between adjacent rings. We discuss these lattices in detail in Section \ref{sec:lattices}.

Regardless of the choice of coupling, networks with hundreds to thousands of oscillators are well beyond the scale of differentiating neural networks previously studied. This section builds the formalism describing the simulations we use to study large networks of ring oscillators.


\subsection{Networks of Differentiating Neurons}
\label{sec:nv-nets-intro}

A network of $n$ differentiating neurons is described by the system of differential equations $\dot{v}(t) = -u(t) + v(t)$ where $v\in \R^n$ is the vector of capacitor voltages, and $u \in \{0,1\}^n$ are the (binary) input voltages to each neuron. The input depends on the (binary) output voltages $y \in \{0,1\}^n$ of the neurons, through a binary function $F_u : \{0,1\}^n \to \{0,1\}^n$. The output voltages $y_i(t)$ depend on the time-derivative of $v_i(t)$ through a function $F_y : \R \to \{0,1\}$. Here, we use $F_y$ describing an inverting Schmitt trigger logic gate, which has a form short-term memory called \textit{hysteresis}. In particular, changes in output are triggered by the capacitor voltages crossing two thresholds $0 < \vthl < \vthh < 1$:
\begin{equation}
y_i(t) = F_y\big(\dot{v}_i(t)\big) =  \begin{cases}
0, & \tau\dot{v}_i(t) \ge \vthh \\
0, & \tau\dot{v}_i(t) \in [\vthl, \vthh) \ \textrm{and} \ y_i(t^{-}) = 0 \\
1, & \textrm{otherwise}
\end{cases}
\end{equation}
where $y_i(t^-)$ denotes the output of neuron $i$ immediately prior to its capacitor voltage $v_i(t)$ crossing one of the thresholds.

The binary nature of the inputs and outputs of the neurons gives rise to a natural \textit{event-based} description of the system dynamics where \textit{events} are defined as changes in output of any neuron in the network. Between these changes in output, the input to each neuron is also fixed, so the time-dependence of the capacitor voltage has a closed-form solution. This property allows one to compute the time between events using only elementary functions of the neuronal inputs and capacitor voltages.

Formally, the capacitor voltages evolve in a continuous and piecewise-smooth manner, with discontinuities in $\dot{v}$ occurring due to changes in output. Letting $t_1, t_2, ...$ denote the times at which these discontinuities occur, denote by $y_i(t_k^-)$ and $y_i(t_k^+)$ the output of neuron $i$ immediately before and after the event at time $t_k$, respectively. Disregarding the relationship between $y_i(t_k^-)$ and $y_i(t_k^+)$ for the moment, the dynamics of the network on the interval $T_k = (t_{k}, t_{k+1})$ follows the equations
\begin{align}
\label{eqn:time-stepping-eqns-1}
y_i(t) &= y_i(t_k^+) \\
\label{eqn:time-stepping-eqns-2}
v(t) &= u(t_k^+) - \tau \dot{v}_i(t) \\ 
\label{eqn:time-stepping-eqns-3}
\tau \dot{v}_i(t) &= \big[u(t_k^+) - v(t_k^+)\big] e^{-(t-t_k)/\tau}
\end{align}
due to the constant output $y(t_k^+)$ and hence constant input $u(t) = F_u\big(y(t)\big)$. Note that the constant output on each interval means we have the equivalence $y(t_k^+) = y(t_{k+1}^-)$, i.e. the neuronal outputs immediately after the $k$-th event are same as they are immediately before the $(k+1)$-st event.

The sequence of event times $\{t_k\}$ can be computed iteratively from the initial condition $\big(v(t_0), y(t_0)\big)$ and the choice of activation function. Not all initial conditions are valid; indeed, the pair $\big(v(t_0), y(t_0)\big)$ must satisfy the consistency relationships 
\begin{equation}
y_i(t_0) = F_y\big(\dot{v}_t(t_0)\big) = F_y\bigg(\frac{1}{\tau}(u_i(t_0) - v_i(t_0))\bigg)   
\end{equation}
where $u(t_0) = F_u\big(y(t_0)\big)$ is the initial vector of input voltages to the neurons. 

Starting from a valid state, all capacitor voltages $v_i(t)$ are either increasing or decreasing over the interval $T_0 = [t_0,t_1)$ prior to the first event. When using the Schmitt trigger to determine neuronal output, a change in output for neuron $i$ only occurs either when $\tau \dot{v}_i$ crosses the threshold $\vthl$ from above or when it crosses $\vthh$ from below. There are three cases to consider in terms of changes in output for such networks:
\begin{enumerate}[leftmargin=*, label=(\arabic*)]
\item \textbf{Autonomous transition 0 $\to$ 1}

\textit{Condition}: $\tau \dot{v}_i < \vthl$, $y_i(t^-) = 0$, and $u_i(t^-) = u_i(t^+)$

This scenario occurs when a neuron stops firing (switches from output 0 to 1) autonomously due to $\tau \dot{v}_i$ continuously dropping below the lower threshold.

\item \textbf{Input-driven transition 0 $\to$ 1}

\textit{Condition}: $y_i(t_-) = 0$, $u_i(t^-) = 1$ and $u_i(t^+) = 0$ 

This scenario occurs when a neuron's input changes from 1 to 0, which causes $\tau \dot{v}_i$ to change discontinuously from a positive value to a negative value. This puts $\tau \dot{v}_i $ below the lower thresholds $\vthl \ge 0$, so the neuron switches output to $y_i(t^+) = 0$.

\item \textbf{Input-driven transition 1 $\to$ 0}

\textit{Condition}: $\tau \dot{v}_i(t) \ge \vthh$, $u_i(t^-) = 0$ and $u_i(t^+) = 1$ 

This scenario occurs when a neuron's input changes from 0 to 1, which causes $\tau \dot{v}_i$ to change discontinuously from a negative value to a positive value. If the capacitor $v_i$ discharged enough previously, then $\tau \dot{v}_i$ will exceed $\vthh$ and the neuron changes output to $y_i(t^+) = 0$.
 
\end{enumerate}
Notice how the latter two cases \textit{are conditioned on another change of output event}. It follows that the first change of output event is necessarily of the first type, namely a neuron stops firing autonomously simply because its capacitor voltage stopped changing rapidly enough. We can therefore find the time $t_1$ of the first event by computing the smallest time at which one of the initially firing neurons satisfies $\tau v_i(t) = \vthl$:
\begin{equation}
    \label{eqn:next-event-time}
    t_1 = \min_{i \ s.t. \ y_i(t_0) = 0} \enskip  - \tau \ln \bigg(\frac{\vthl}{u_i(t_0) - v_i(t_0)}\bigg) 
\end{equation}
The neurons that will cause the first event are those achieving the minimum (there could be multiple such neurons).

The capacitor voltages of the neurons at the first event can be computed explicitly once we've found $t_1$ using Eqn. \ref{eqn:time-stepping-eqns-1}-\ref{eqn:time-stepping-eqns-3}.

Determining the neuronal outputs $y(t_1^+)$ immediately after this change in output event requires some care because changes in a single neuron's output can induce a cascade of changes in neuronal output throughout the network. For instance, when a neuron switches from output 0 to 1, one of its children may immediately switch from output 0 to 1, potentially inducing even further changes in neuronal activity in the network. These are the ``input-driven`` transitions mentioned previously.

We've modeled such cascading effects due to changes in output as occurring instantaneously, though in hardware there will be a small transmission delay. This choice means we must explicitly compute the result of such cascades when updating the neuronal output vector $y(t_1^-)$ to $y(t_1^+)$ due to a change in output at time $t$. We use the Algorithm \ref{alg:time-stepping} to propagate the state of the lattice forward in time from one autonomous output transition to the next.

\begin{algorithm}[!hb]
\DontPrintSemicolon
  \KwInput{
    \newline $t_0$, current simulation time \newline $v(t_0)$, capacitor voltage vector  \newline $y(t_0)$, output voltage vector \newline $t_e$, next event time \newline $\mathcal{E}$, indices of the neurons causing the event
  }
  \KwOutput{
    \newline $v(t_e)$, capacitor voltage vector at time $t_e$ \newline $y(t_e)$, output voltage vector at time $t_e$ \newline $t_e'$, new next event time \newline $\mathcal{E}'$, indices of the neurons causing the new event 
  }
  \tcp{Step 1: Update the capacitor voltages of all neurons given the current outputs.}   
  \For{each neuron index $i$}{
      $v_i(t_e) \gets u_i(t_0) + \big(v_i(t_0) - u_i(t_0)\big) e^{-(t_e - t_0) / \tau}$
  }
  \tcp{Step 2: Record the autonomous transitions.}
  \For{neuron index $i$}{
    \If{$i \in \mathcal{E}$}{
        $y_i(t_e) \gets 1$
    } 
    \Else{
        $y_i(t_e) \gets y_i(t_0)$
    }
  }
  \tcp{Step 3: Propagate these changes in output.}
  $\mathcal{U} \gets \bigcup_{i \in \mathcal{E}}\{\text{children of node } i\}$  \;
  \While{$\mathcal{U}$ is not empty}{
        $i \gets$ pop$(\mathcal{U})$ \;
        $u_i(t_e) \gets F_u\big(y(t_e)\big)_i$ \;
        $\dot{v}_i(t_e) \gets u_i(t_e) - v_i(t_e)$ \;
        \tcp{Input-driven transition 0 $\to$ 1}
        \If{$y_i(t_0) = 0, u_i(t_0) = 1$, and $u_i(t_e) = 0$}{
            $y_i(t_e) = 1$ \;
            $\mathcal{U} \gets \mathcal{U} \cup \{ \text{children of node} \  i\}$
        }
        \tcp{Input-driven transition 1 $\to$ 0}
        \ElseIf{ $\dot{v}_i(t_e) \ge \vthh$, $u_i(t_0) = 0$, and $u_i(t_e) = 1$}{
            $y_i(t_e) = 0$ \;
            $\mathcal{U} \gets \mathcal{U} \cup \{ \text{children of node} \  i\}$
        }
  }
  \tcp{Step 4: Compute the next autonomous transition.} 
  $t_e' \gets \infty$ \;
  $\mathcal{E}' \gets \{\}$ \;
  \For{neuron index $i$ s.t. $y_i(t_e) = 0$}{
    $u_i(t_e) \gets F_u \left( y(t_e)\right)_i$ \;
    $t' \gets -\tau \ln \left( \vthl / (u_i(t_e) - v_i(t_e)) \right)$ \;
    \If{$t' < t_e'$}{
        $t_e' \gets t_e$ \;
        $\mathcal{E}' \gets \{i\}$ \;
    }
    \ElseIf{$t' = t_e'$}{
        $\mathcal{E}' \gets \mathcal{E}' \cup \{i\}$ \;
    }
  }
  \Return{$v(t_e), \ y(t_e), \ t_e', \ \mathcal{E}'$}
\caption{Time-Stepping Algorithm}
\label{alg:time-stepping}
\end{algorithm}

To compute the lattice state at time $t > 0$ starting from an initial condition $(v(0), y(0))$, we use Eqn. \ref{eqn:next-event-time} to compute the time $t_1$ of the first event and the associated event nodes $\mathcal{E}$. The time-stepping routine is then called iteratively, stepping the network state forward from one event to the next, until an event occurs after time $t$. Letting $\left(v(t_{n}), y(t_{n})\right)$ and $\left(v(t_{n+1}), y(t_{n+1})\right)$ denote states at the penultimate and final events, respectively, the state at time $t$ follows from Eqns \ref{eqn:time-stepping-eqns-1}-\ref{eqn:time-stepping-eqns-2}:
\begin{align*}
    u_i(t_n) &= F_u\left(y(t_n)\right)_i \\
    v_i(t) &= u_i(t_n) + \left[v_i(t_{n}) - u_i(t_n) \right] e^{-(t-t_n)/\tau} \\
    y_i(t) &= y_i(t_{n})
\end{align*}

\textbf{Note} that modeling the differentiating neurons as having binary outputs and making discontinuous changes in output \textit{prevents the existence of adjacent firing neurons}. This choice is the source of the restricted output state-space mentioned in Section \ref{sec:background-ring-oscillators}

\textbf{A word of warning} ---  When simulating networks with odd-length cycles, Step 3 of Alg. \ref{alg:time-stepping} can sometimes enter an infinite loop when all neurons within the ring are capable of firing. The phenomenon arises from our choice to model the logical outputs of the neurons as being able to transition between 0 and 1 instantaneously, whereas in hardware this is a small amount of time required for this turnover to occur. Physical circuits with odd-length cycles of neurons are capable of oscillating with frequencies on the order of this turnover time. Such oscillations are costly to simulate, so this paper focuses exclusively on networks with exclusively even-length cycles, where empirically these rapid oscillations do not arise during our simulations.

\begin{figure*}[ht]
    \includegraphics[width=1.8\columnwidth]{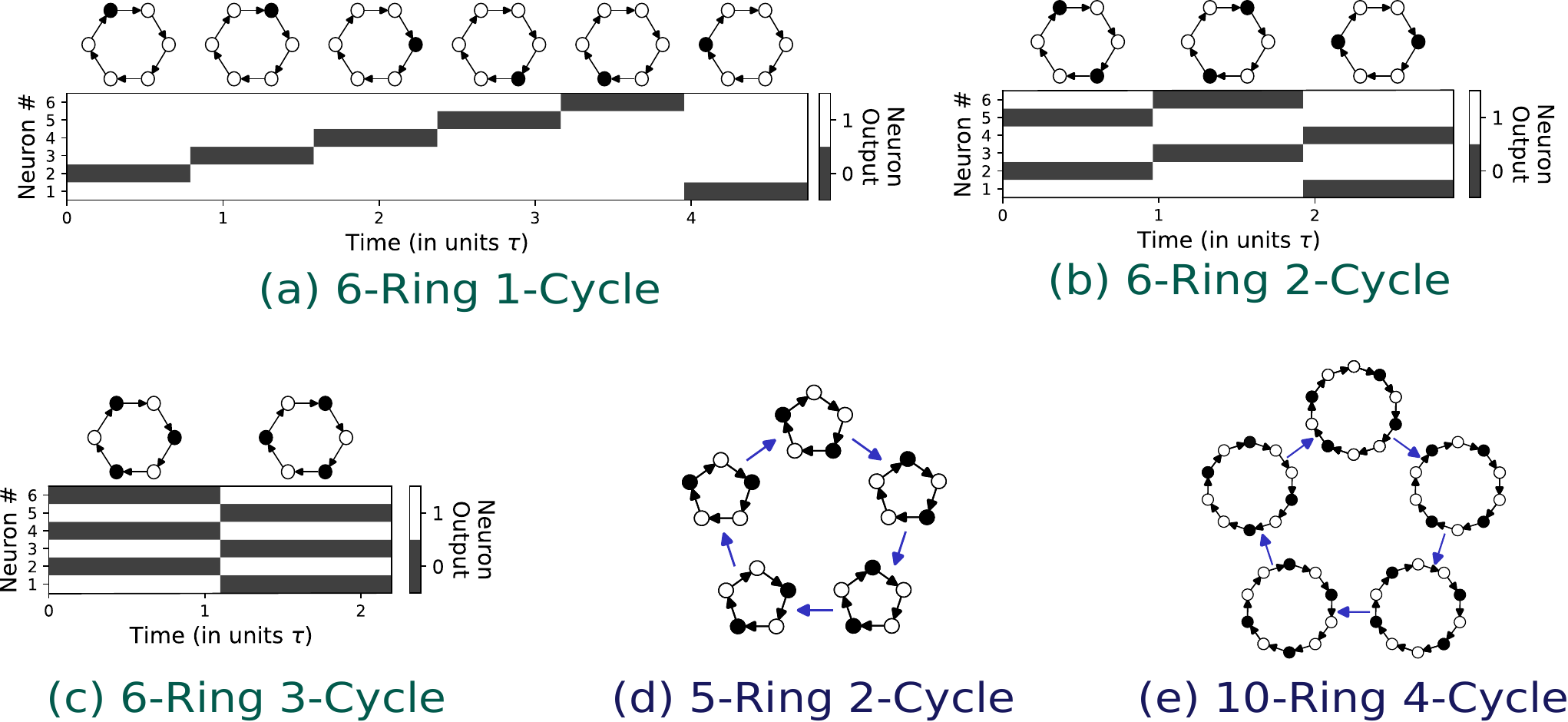}
    \caption{
    \label{fig:ring-cycles}
    \textbf{Periodic Orbits of the Ring Oscillators} \ \ 
	\textbf{(a) --- (c)} show the $k=1,2,3$ pulse cycles for the $n=6$-ring oscillator. Some pulses are synchronized in their movement for $k=2$ and $k=3$, with this property arising from the the ring size being a multiple of the pulse counts. The periods of the cycles are not rational multiple of each other, and they depend on the threshold lower threshold $v_{\textrm{thl}}$ (c.f. S\ref{sec-oscillator-period} for details). 
     \textbf{(d)} The $k=2$ cycle of the $n=5$ ring oscillator is different from the 2-cycle of the 6-ring because the pulses are not synchronized in movement; instead, they alternate moving from neuron to neuron. \textbf{(e)} The $k=4$ cycle on the $n=10$ ring exhibits both pulses that move together and pulses that move separately. Each neuron oscillates on and off with duty cycle $4/10 = 2/5$. Neurons a distance of 5 apart with therefore be in phase because the oscillation period of each neuron is an integer multiple of $(2/5)P$. In this way, the $4$-cycle of the $10$-ring behaves like two copies of the $2$-cycle for the $5$-ring. This result seems to generalize: for integer $f$, the $fk$-cycle of a $fn$-ring behaves like $f$ copies of the $k$-cycle of an $n$-ring. 
}
\end{figure*}

%
%
\section{Periodic Orbits of Differentiating Ring Oscillators}
\label{sec:ring-oscillator-periodic-orbits}

\begin{figure}[!t]
    \includegraphics[width=0.95\columnwidth]{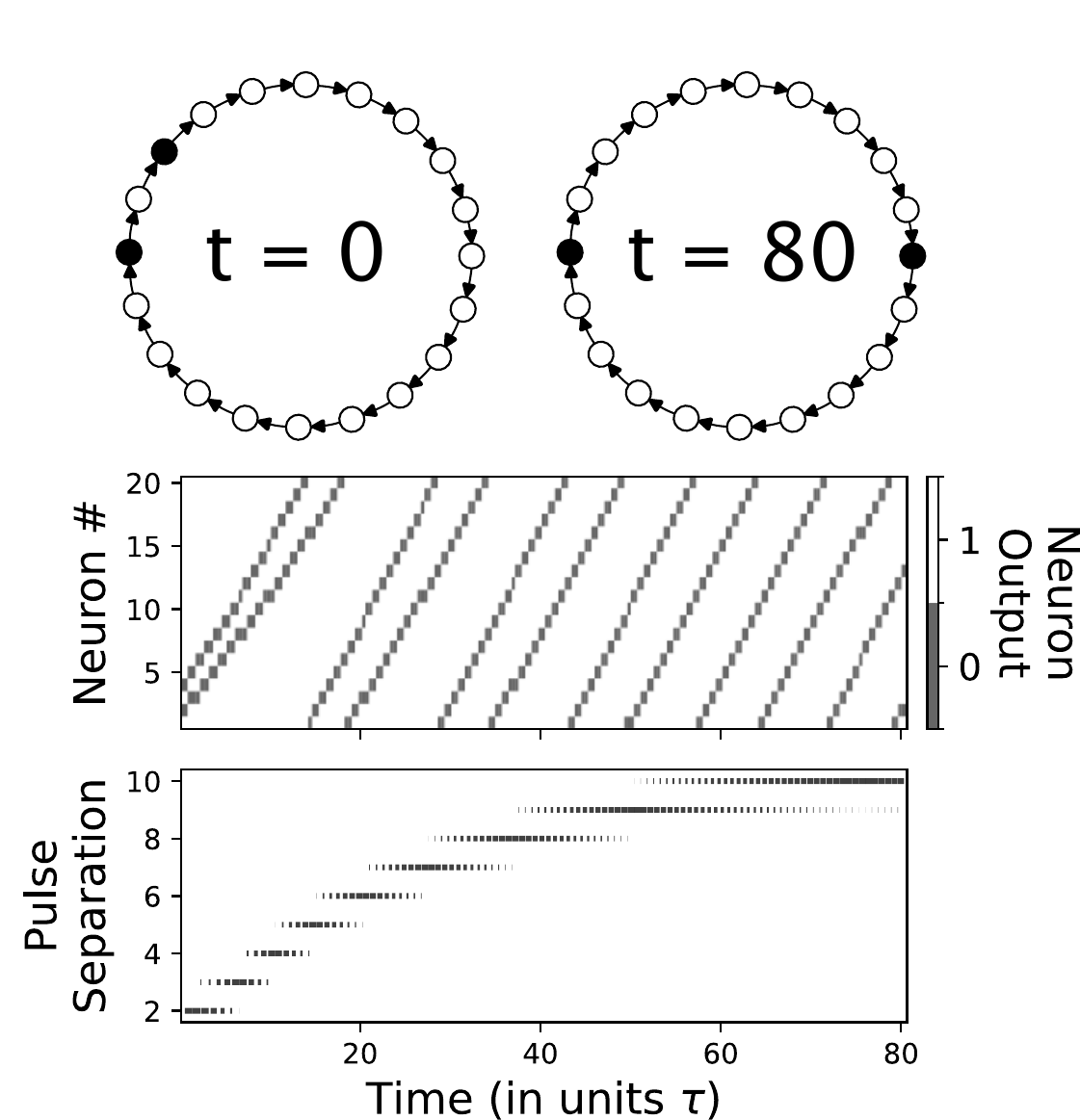}
    \caption{
    \label{fig:pulse-separation}
    \textbf{Pulse separation} shown as a 20-ring oscillator converges to its 2-cycle. The ring starts in a state with two firing neurons that are close to each other in the ring. The separation (number of neurons) between the two pulses increases over time, growing exponentially from 2 neurons to 10 neurons as the ring converges to its 2-pulse periodic orbit. Pulse separation is not strictly increasing in time, but we see fluctuations between consecutive integer values with the larger integer becoming more probable over time.}
\end{figure}

Before presenting our experiments with the ring oscillator lattices, we will first discuss the nature of the periodic orbits of the ring oscillators. Characterization of these orbits will be necessary to perform \textit{phase reduction}, which we'll use to study networks of coupled ring oscillators in the next section.

 This section presents the results of these simulations and characterizes the set of stable periodic orbits of the oscillators.

\subsection{Stable Periodic Orbits}
\label{sec:ring-oscillator-periodic-orbits-characterization}

Past research \citep{Rietman03} suggests that the number of periodic orbits grows exponentially with the number of neurons in the ring. However, in our experiments, we find that most of these periodic orbits are not stable; in fact, only $\lfloor n/2 \rfloor$ of these orbits are stable attractors. In the remainder of this section, we present numerical results to support this claim and provide characterization of the observed stable cycles.

We found that all $n$-ring oscillators eventually converged to one of $\lfloor n/2 \rfloor$ unique periodic cycles, one for each possible pulse count. One way of characterizing these limit cycles is by the proportion of time each neuron in the ring is firing/inactive, referred to herein as the neuron's \textit{duty cycle}. Given an $n$-ring harboring $k$ pulses, we have observed that the ring will converge such that each neuron's output switches between 0 and 1 with duty cycle $k / n$ and with the same period. This characterization suffices to uniquely define the cycle due to the ring connectivity. If $P$ is this period, each neuron is $(k/n) P$ out of phase with its parent as each neuron starts firing immediately after its parent stops firing; hence, a pair of neurons that have $d$ other neurons between them will exhibit a $(dk/n)P$ phase difference. As such, the signals output by the neurons in the ring are all equivalent under time translation in units of $(k/n)P$. Fig. \ref{fig:ring-cycles} shows the three stable cycles of a $6$-ring. 

We find that long-range synchronization of neurons can occur in rings where $n$ is not prime due to the rational differences in neural output phase. For instance, if $g = \gcd(n,k)$, then $\frac{n}{g} \big(\frac{k}{n}\big) P \in P \mathbb{Z}$. Equivalently, in the $k$-cycle of an $n$-ring, neurons which are a distance $n / \gcd(n,k)$ apart will be an integer multiple of $P$ units out of phases, which implies they will be synchronized as $P$ is the oscillation period of each neuron. In this way, long-range synchronization occurs when $n$ has large factors. An example of synchronization can be seen in the limit cycles of the $6$-ring in Fig. \ref{fig:ring-cycles}b and \ref{fig:ring-cycles}c. In former, we see that the two pulses move simultaneously because they are a distance 3 apart and $6 / \gcd(6,2) = 3$. On the other hand, the two pulses do no move simultaneously for the 2-cycle of the 5-ring as shown in Fig. \ref{fig:ring-cycles}d.

Another characterization of the periodic orbits of the $n$-ring oscillator is that pulses tend to separate spatially over time as if they were experiencing a repulsive force. As an example, consider the equivalence classes of a 6-ring oscillator output states with two pulses. The pulses in the state $100100$ are further apart spread out than the pulses in state $101000$, and we find that the 2-cycle of the 6-ring comprises all states isomorphic under rotation to 100100. Another example of this pulse separation can be seen in Fig. \ref{fig:pulse-separation} showing the convergence of a 20-ring oscillator to its 2-cycle.

\subsection{Oscillation Period}
\label{sec-oscillator-period}

From this characterization of the periodic orbits of the ring oscillator, we can derive an implicit equation for the \textit{period} of these cycles. During the orbit each neuron produces the same output, just shifting in time, so it suffices to consider only a single neuron to compute this value.

Let $P$ denote the period of the $k$ cycle of the $n$-ring oscillator (in units $\tau$), and define the start of the cycle as the point when the first neuron has just stopped firing. The neuron undergoes three phases during the cycle: it fires for $(k/n) P$ units of time, its parent neuron fires for $(k/n) P$ units of time, and there are $(1 - 2k/n) P$ units of time in between when both the neuron and its parent are dormant.
\begin{enumerate}[leftmargin=*]
\itemsep0em
\item The neuron stops firing when it's voltage takes value $v_0 = 1 - v_\textrm{thl}$, and it must wait $(1-2k/n)P$ units of time before the next pulse reaches it again. During this time its input is 1 V because the parent is dormant, so its capacitor voltage charges to up
\begin{equation}
v_1 = 1 + (v_0 - 1) e^{-(1-2k/n)P} = 1 - v_\textrm{thl}e^{-(1-2k/n)P}
\end{equation}
\item The neuron then receives input of 0 V for $(k/n) P$ units of time as its parent is firing, so its voltage discharges from $v_1$ to
\begin{equation}
v_2 = v_1 e^{-(k/n)P} = e^{-(k/n)P} - v_\textrm{thl} e^{-(1-k/n)P}
\end{equation}
\item Finally, the neuron itself firing for another $(k/n)P$ units of time before its capacitor voltage returns to $1 - v_\textrm{thl}$. During this time the parent is dormant, so the neuron's input is 1 V. It's capacitor voltage therefore changes from $v_2$ to
\begin{equation}
v_3 = 1 + (v_2 - 1)e^{-(k/n)P} = 1 + e^{-(2k/n)P} - e^{-(k/n)P} - v_\textrm{thl} e^{-P}
\end{equation}
\end{enumerate}
Equating $v_3$ with $1-v_\textrm{thl}$ and simplifying, the equations above yield the governing equation for the period of the cycle:
\begin{equation}
e^{-(2k/n)P} - e^{-(k/n)P} - v_\textrm{thl} e^{-P} + v_\textrm{thl} = 0
\end{equation}
or equivalently
\begin{equation}
p_{nk}(x) = v_\textrm{thl} x^n - x^{2k} + x^k - v_\textrm{thl} = 0 
\label{eqn:period-polynomial}
\end{equation}
where $x = \exp(-P/n)$. This yields a degree $n$ polynomial equation from whose roots we can compute the possible periods of the $k$-cycle of the $n$ ring oscillator. 

With $P > 0$, the roots $x = \exp(-P/n)$ of meaningful to the ring oscillator are those that fall within the unit interval $[0,1]$. We've found empirically that there are at most two of such roots to $p_{nk}$ depending on the value of the threshold $v_\textrm{thl}$. The larger of these roots appears to coincide with an unstable periodic orbit, however, so the ratio $k/n$ uniquely defines the period of the stable orbits.

%
%
\section{Ring Oscillator Lattices}
\label{sec:lattices}

\begin{figure*}
    \includegraphics[width=1.9\columnwidth]{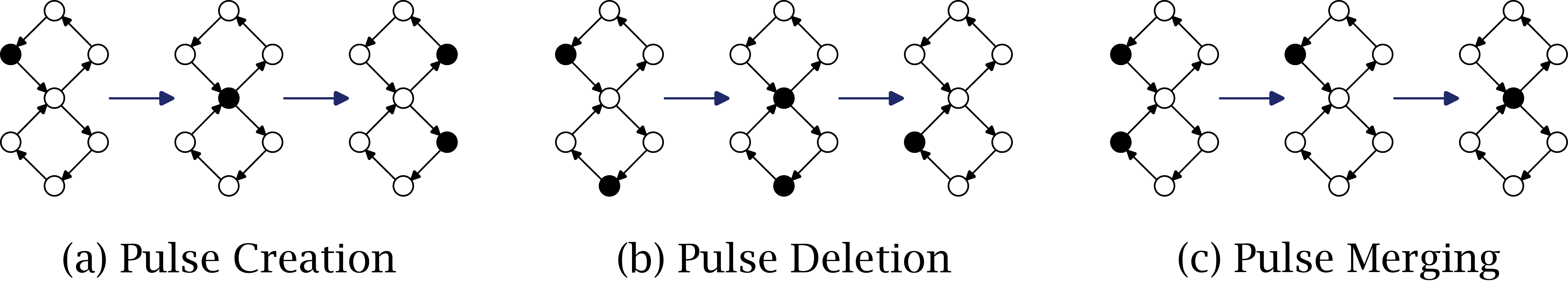}
    \caption{
    \label{fig:local-interactions}
    \textbf{Local Interactions between Rings} \textbf{(a) Pulse Creation} can occur when a ring has more active neurons than an adjacent ring. Here, the top 4-ring initially has one active neuron whereas the bottom ring has none. The pulse in the top ring is transferred to the central shared neuron, and this pulse is subsequently transferred to one neuron in each ring, increases the total number of pulses in the network. 
	\textbf{(b) Pulse Deletion} \ Both the top and bottom ring both initially have one pulse, but they are in different locations in the rings. The top ring is the first to transfers its pulse first to the shared neuron. The movement of the other pulse in the bottom ring stops the shared neuron from firing, however, because a neuron cannot fire when any of its parents are firing.
	\textbf{(c) Pulse Merging} can occur at shared neurons that receive inputs from neurons in multiple rings. If the parents of such neuron are not synchronized, the transmission of their pulses to the shared neuron does not happen until both parents stop firing. This affects the timing of subsequent neuron activity in both rings.
}
\end{figure*}

\begin{figure}
    \includegraphics[width=0.8\columnwidth]{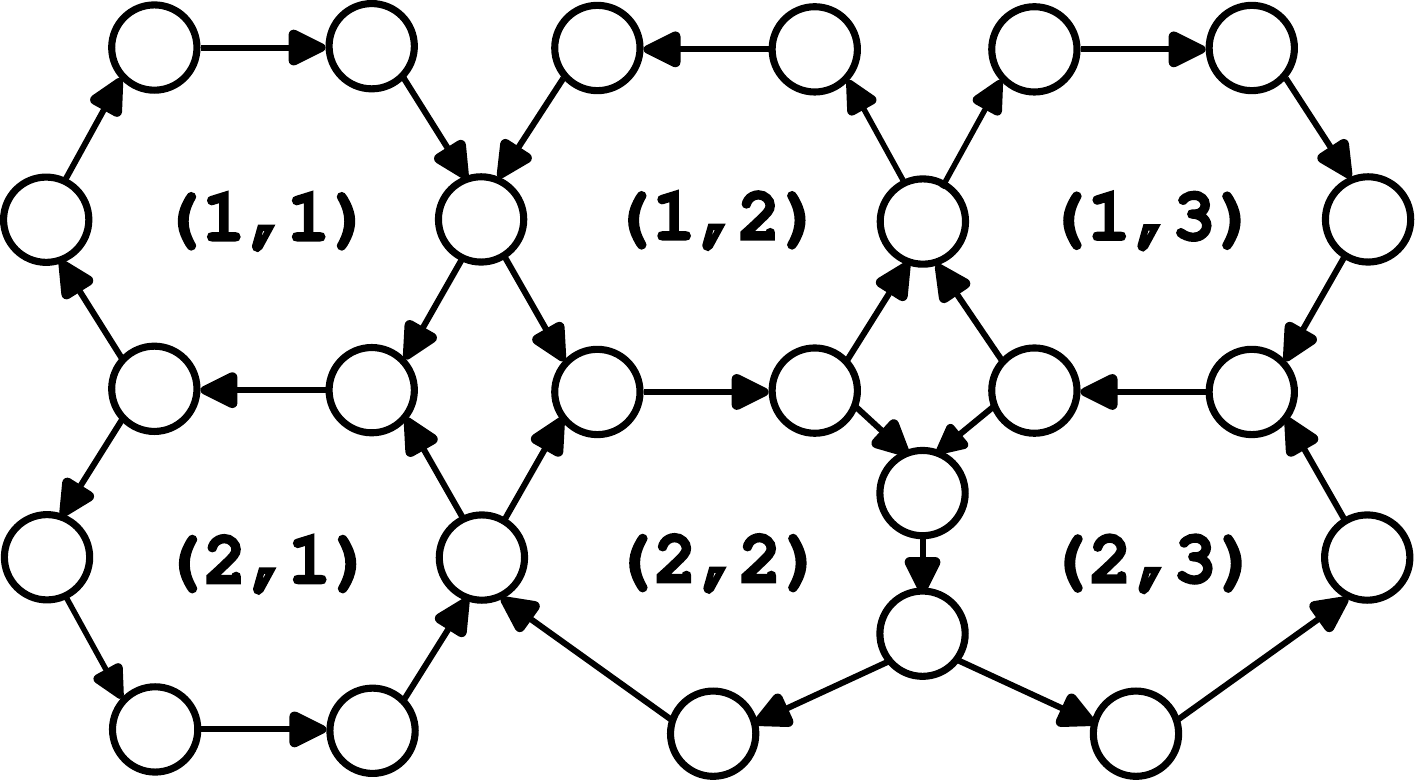}
    \caption{
    \label{fig:basic-lattice-example} 
    \textbf{Example lattice of 6-ring oscillators} Note how each ring shares an edge with one of its neighbors, so the oscillation direction flips (CW $\leftrightarrow$ CCW) from ring to ring.
    }
\end{figure}

We next investigate the dynamics of ring oscillator lattices from the perspective of \textit{oscillator synchronization}. In the previous section, we saw that each non-interacting ring converges to a periodic orbit characterized by the number of pulses in the ring. When the rings are coupled into lattices, pulses in one ring can be added, removed, and edited by the activity of other rings, and this interaction disrupts convergence to their periodic orbits.

We find that the periodic orbits of the individual rings are not stable for arbitrary lattices of rings. In this section, we therefore focus on a class of \textit{homogeneous} lattices that support global periodic orbits in which all rings have synchronized to the same periodic orbit. We then use \textit{phase reduction} analysis to define synchronization \textit{correlation lengths}. These length scales elucidate the transient dynamics that emerge as the ring oscillators synchronize with each other starting from a random initial state.

\subsection{Coupling and Lattice Construction}

The lattices studied in this section are constructed by placing a ring oscillator at each site of a rectangular lattice. Rings are subsequently coupled with each other by \textit{sharing} neurons between adjacent rings, such as in Fig \ref{fig:basic-lattice-example}. Note how the oscillation direction (clockwise vs. counter-clockwise) is opposite for any pair of adjacent rings. We made this choice to ensure that there are no bidirectional edges in the network when adjacent rings share multiple neurons.

The choice of coupling creates situations where certain neurons receive input from multiple parent nodes. In these cases, we define the input voltage for the neuron as 0 if \textit{any} of the parents are firing, and 1 otherwise. This choice is equivalent to placing an AND gate before the capacitor through which all inputs are connected. When a neuron has multiple children, both children receive the same input equal to the exact output of the parent neuron. 

The overall connectivity of a lattice is encoded by five variables per ring --- the ring size and how the ring shares neurons with its neighbors. We start with an $N \times M$ rectangular lattices and place a ring with $N_{ij}$-neurons at each site. The lattice is constructed so that the ring at site $(i,j)$ shares $L_{ij}$ nodes with the ring to its right (site $(i-1,j)$), $T_{ij}$ nodes with the ring above it (site $(i,j-1)$), $R_{ij}$ nodes with the ring to it's right (site $(i,j+1)$), and $B_{ij}$ nodes with the ring below is (site $(i+1,j)$). For instance, in Fig. \ref{fig:basic-lattice-example}a, the ring at site (1,2) has $L_{12} = R_{12} =  1$ and $B_{12} = 2$, whereas ring (2,2) has $L_{22} = 1$ and $T_{22} = B_{22} = 2$. 

For simplicity, all neurons are shared between some pair of rings, e.g. we require $N_{ij} = L_{ij} + T_{ij} + R_{ij} + B_{ij}$. Note also that these variables are not all independent. For instance, the ring at lattice site $(i,j)$ shares $R_{ij}$ neurons with its neighbor $(i,j+1)$. These shared neurons are the \textit{left} neurons of the ring at site $(0,1)$, so this forces $R_{ij} = L_{i,j+1}$.

\begin{figure*}
   \includegraphics[width=1.7\columnwidth]{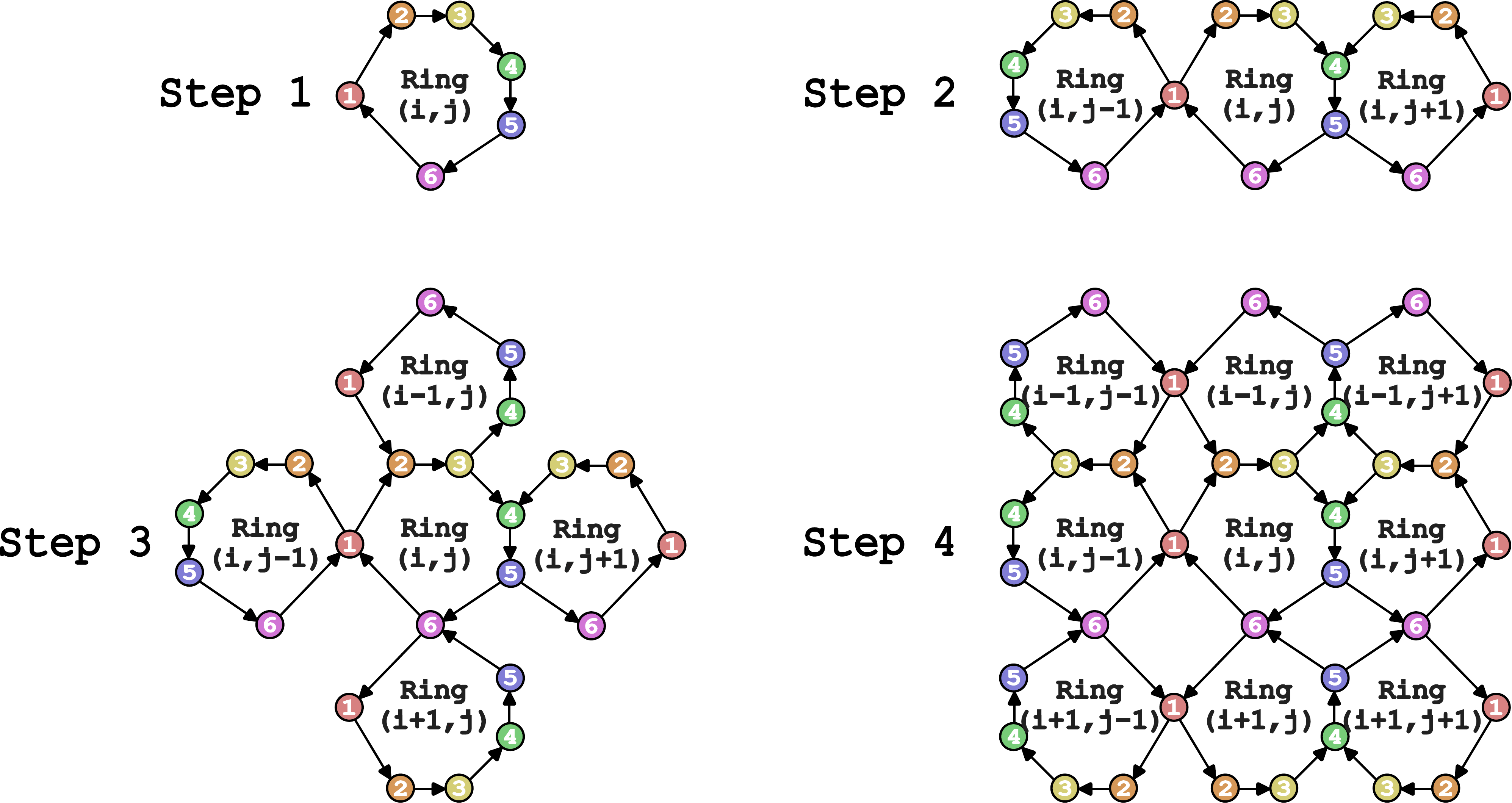}
   \caption{
     \label{fig:hom-lattice-construction}
     \textbf{Example homogeneous lattice construction} \ 
     \textbf{Step 1}: Start with the connectivity template $L_{ij} = 1, T_{ij} = 2, R_{ij} = 2$, and $B_{ij} = 1$ for the ring at site $(i,j)$. The nodes in the template are colored to show how the construction algorithm maintains homogeneity in the network. \ \textbf{Step 2} This template is reflected across the left and right edges of site $(i,j)$ to make the rings at site $(i,j\pm 1)$.
     \textbf{Step 3}: The template for ring $(i,j)$ is reflected across its top and bottom edges to generate the templates for rings $(i\pm 1, j)$.
     \textbf{Step 4}: The connectivity for the rings constructed in Steps 2 and 3 are used to define the connectivity in rings $(i-1,j\pm 1)$ and $(i+1,j\pm 1)$. 
     Notice how the homogeneous property of the lattice is still maintained --- any cycle of 6 nodes cycles through the colors red, orange, yellow, green, blue, and purple in the same order.
    }
\end{figure*}

We saw in \ref{sec:ring-oscillator-periodic-orbits} that the long-term behavior of an autonomous ring oscillator can be characterized by the number of pulses in the ring. Inter-ring coupling enables the pulses within the rings to interact, resulting in the \textit{creation}, \textit{deletion}, and \textit{merging} of pulses moving throughout the network like in Fig. \ref{fig:local-interactions}. These local interactions affect the pulses in the rings and disrupt the convergence of the rings to periodic orbits. However, the influence of these local interactions on the global dynamics of the lattice is not immediately evident. In the later subsections, we employ numerical simulations to study the transient and long-term behavior of these oscillator lattices.

\subsection{Homogeneous Lattices}

We study a special case of oscillator networks that support global periodic orbits where all rings have synchronized to the same periodic orbit. 

Let $G = (V,E)$ denote the directed graph describing a differentiating neuron network, where there is an neuron at each vertex $v \in V$ and there is an edge $(u,v) \in E$ if neuron $v$ receives input from neuron $u$. We say that this network is $k$-\textit{homogeneous} if there exists a coloring function $c : V \to \{1, 2,...,k\}$ such that nodes assigned color $i$ receive input only from nodes with color $i-1 \bmod{k}$. This is to say that for all $(u,v) \in E$, $c(v) \equiv c(u) + 1 \bmod{k}$. Intuitively, any simple cycle of nodes in the graph must cycles through all $k$ colors periodically. This restriction ensures all cycles in the graph have length $mk$ for some integer $m > 1$, which we know from Section \ref{sec:ring-oscillator-periodic-orbits} independently supports a periodic orbit with $k$ pulses.

We use the following iterative algorithm using reflections to ensure homogeneity holds in a lattice. Refer to Fig. \ref{fig:hom-lattice-construction} for an example of the process.
\begin{enumerate}[wide=0pt, leftmargin=15pt, labelwidth=10pt, align=left]
\item Fix a ring size $N$ and connectivity $L_{ij} = L$, $T_{ij} = T$, $R_{ij} = R$, and $B_{ij} = B$ site $(i,j)$. 
\item Set the connectivity for the ring at lattice sites $(i,j\pm 1)$ as $L_{i,j\pm 1} = R$, $T_{i,j\pm 1} = T$, $R_{i,j\pm 1} = L$, and $B_{i,j\pm 1} = B$. This construction makes the rings at site $(i,j\pm 1)$ looks like reflections of the ring at site $(i,j)$ across their shared edge. 
\item Set the connectivity for the ring at lattice sites $(i\pm 1,j\pm 1)$ as $L_{i\pm 1,j} = L$, $T_{i\pm 1,j} = B$, $R_{i\pm 1,j} = L$, and $B_{i\pm 1,j} = T$. This choice similarly makes the ring at site $(i\pm 1,i)$ looks like a reflection of the ring at site $(i,j)$ across their shared edge. 
\item Iteratively apply steps 2 and 3 to determine the connectivity of the remaining rings in the lattice. As an example, the ring at site $(i+1,j+1)$ will have $L_{i+1,j+1} = R$, $T_{i+1,j+1} = B$, $R_{i+1,j+1} = L$, and $B_{i+1,j+1} = T$. 
\end{enumerate}

The construction using reflections ensures the existence of global limit cycles in which each ring oscillates at one of its natural frequencies. An arbitrary lattice is not guaranteed to have this property even if each ring oscillator has the same number of neurons. 

\subsection{Measuring Local Correlations in Oscillator Phase}
\label{sec:lattices-measuring-correlations}

\begin{figure*}[t!]
   \includegraphics[width=1.97\columnwidth]{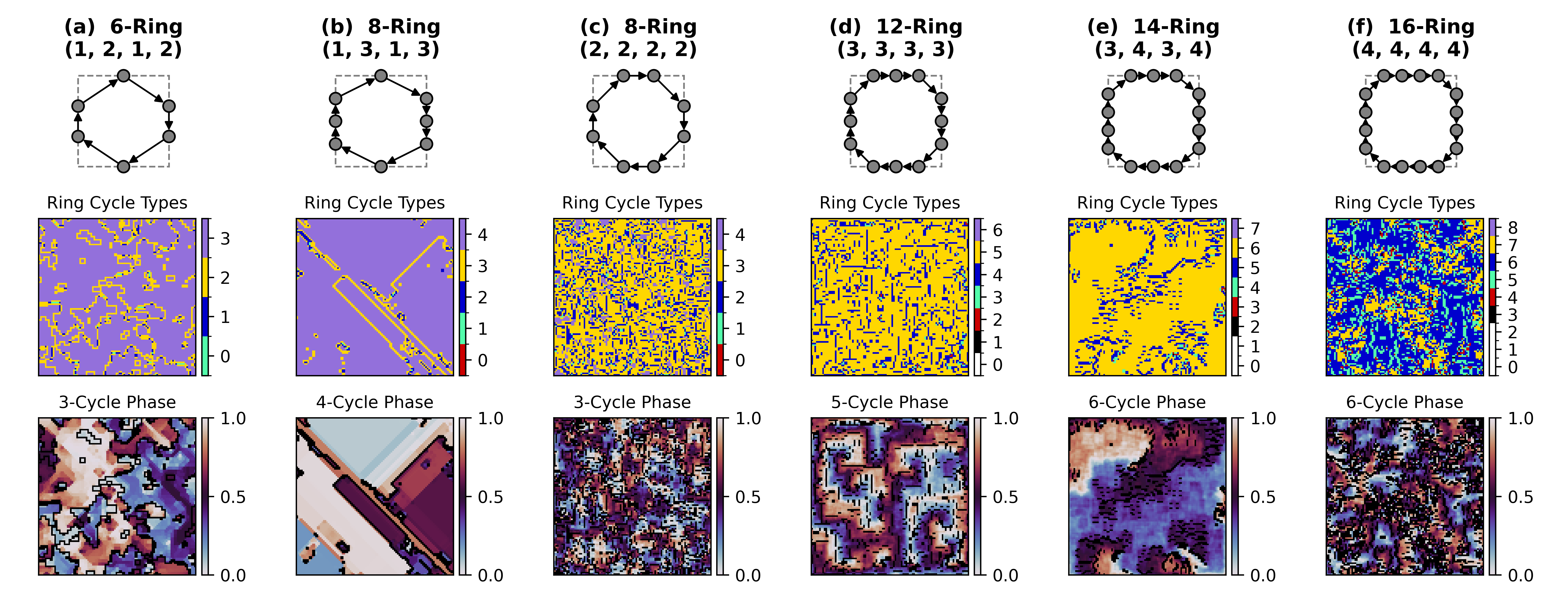}
   \caption{
     \label{fig:phase-patterns}
     \textbf{Transient phase domains} exhibited by 100x100 ring-oscillator lattices. Columns delineate different connectivity template $(T, R, B, L)$ indicated by the top row. Each column shows a snapshot of the cycle types $k_{ij}$ (middle) and associated phases $\theta$ of a specific cycle type (bottom) at time $t = 1000 \tau$, starting from a random initial state.
     In each network, the ring oscillators organize into domains with the same dominant cycle type $k_\textrm{dom}$, with rings in non-dominant cycles forming the domain boundaries.
     The dominant cycle type depends on the the lattice connectivity, not just the ring size $N$, but it does not deviate far from the most saturated cycle $\lfloor N/2 \rfloor$.
     Traversing domain boundaries in the middle plot coincides with rapid changes in phase in the bottom plot, so the adjacent domains are out-of-phase with each other.
     The resulting domains and phase patterns in the network vary greatly in size and shape by ring connectivity, even for rings with the same number of neurons; for instance, compare the 8-ring networks with \textbf{(b)} $(T,R,B,L) = (1, 3, 1, 3)$ and \textbf{(c)} $(T, R, B, L) = (2,2,2,2)$ above. 
     \textbf{(f)} rings with $N=16$ and $(T, R, B, L) = (4,4,4,4)$  are unique out of those simulated in this study, as these networks see the co-existences of \textit{multiple} cycle types forming large phase domains. One can see continuous regions of the network in the 6-cycle (blue) and 7-cycle (gold).
    }
\end{figure*}

In subsequent sections, we present numerical simulation of homogeneous lattices starting from random initial conditions. We find that these networks can indeed settle to their global periodic orbits, but there is an extended transient period as the scale of oscillator synchronization grows. 

To characterize this synchronization numerically during the transient, we extend the technique of oscillator \textbf{phase reduction} (Sec. \ref{sec:background-phase-reduction}) to account for the multiple period orbits accessible to each $n$-ring oscillator.
To each state $\mathbf{x} = (\mathbf{v}, \mathbf{y})$ of the ring, we assign two variables: \textit{the number of pulses} $k$ in period orbit to which the ring converges starting from state $\mathbf{x}$ and the associated \textit{phase} $\theta$ of that oscillation, defined as in Eqn. \ref{eqn:phase-reduction}. More specifically, 
\begin{itemize}[leftmargin=*]
\itemsep0em 
\item For points on the $k$-cycle itself, we can define this phase by defining a mapping between the orbit and the circle. We elect $\theta = 0$ as the point in the limit cycle when the first neuron starts firing, i.e. the point where $v_1$ and $y_1$ changes discontinuously  from $(1-v_\textrm{vthl},1)$ to $(1 - v_\textrm{vthl}, 0)$. 
\item For a point $\mathbf{x} = (\mathbf{v}, \mathbf{y})$ that converges to the $k$-cycle, we assign the associated phase by simulating the time evolution of a ring oscillator starting in state $\mathbf{x}$ to estimate the point on the orbit $\gamma(\mathbf{x}) = \lim_{n\to\infty}\Phi(\mathbf{x},nP)$ from which we assign it phase $(\theta \circ \gamma)(\mathbf{x})$ (See Section \ref{sec:background-phase-reduction} for details).
\end{itemize}
Phase reduction associates a cycle number $k$ and phase $\theta$ with the state of an $n$-ring oscillator. We then characterize the state of an $n$-ring oscillator lattice by these values for the rings at each site, denoted here by $(k_{ij}, \theta_{ij})$ for lattice site $(i,j)$. 

The cycle types and phases can change over time for each ring due to the coupling of the lattice. In particular, the variables $k_{ij}$ and $\theta_{ij}$ can change only when a neuron \textit{external} to ring $(i,j)$ has changed the input to one of the neurons within the ring itself. Comparing this Fig. \ref{fig:local-interactions}, one can see that such an event where a neuron's input is affected be an external node coincides with the three local interactions --- pulse creation, deletion, and merging. 

Through our simulations, we find that these small-scale interactions give rise to complex global dynamics. We will see in the next section that these homogeneous lattices tend toward \textit{local} oscillator synchronization --- the event where $k_{ij}$ and $\theta_{ij}$ are nearly same for all rings in different regions of the network. These phase variables thus allow us to numerically quantify the scale of synchronization in these networks.

\begin{figure*}[t!]
   \includegraphics[width=1.97\columnwidth]{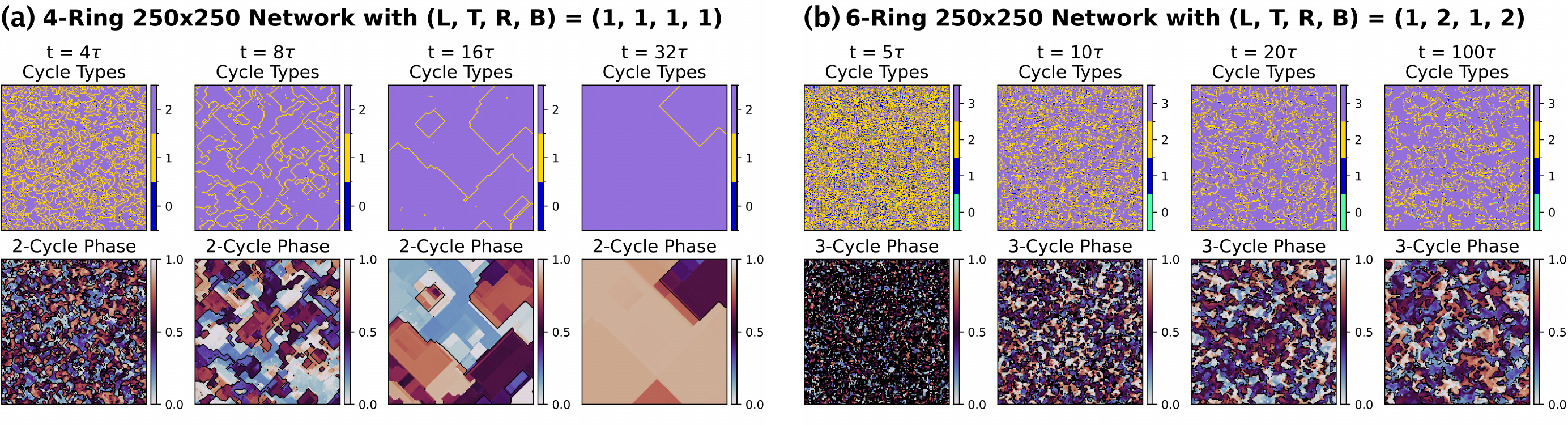}
   \caption{
     \label{fig:example-trajectories}
     \textbf{Cycle type and phase snapshots from small-ring network trajectories}. \textbf{(a)} Example trajectory of a 250x250 4-ring network started from a random initial state, with 30\% of the neurons firing.
     Each column shows a snapshot in time ($t = 4 \tau, 8 \tau, 16 \tau,$ and $32 \tau$) of the cycle types $k_{ij}$ (top) and associated phases $\theta$ of a specific cycle type (bottom).
     At $t = 4\tau$ most rings have a single firing neuron and converge to the 1-cycle (yellow pixels). As time progresses, increasingly many rings saturate to the 2-cycle (purple pixels) and begin forming domains separate by rings in other cycle types. These domains grow in size as the network evolves in time, eventually ending with global synchronization (not shown).
     \textbf{(b)} Example trajectory of a 250x250 6-ring network with $(L,T,R,B) = (1,2,1,2)$, also started from a random initial state with 30\% of the neurons firing. This system similarly forms domains of rings in the fully saturated cycle type ($N/2 = 3$), but these domains never grow to include the entire lattice, even for times on the order of $10^6 \tau$. Visually, the size of the phase domains at time $t=20\tau$ are comparable to those at $t = 100\tau$. 
     Fig. \ref{fig:small-ring-correlations}(a) numerically characterizes this saturation domain size using phase correlation length.
    }
\end{figure*}

To this aim, we use the notion of \textbf{correlation length} \citep{Sethna2021StatMechCorrFunc} from statistical physics to characterize the scale of synchronization in these lattices. We first define a similarity metric on the cycle type-phase pairs $(k,\theta)$ and $(k',\theta')$ for two rings:
\begin{equation}
\langle k,\theta; k',\theta' \rangle = \begin{cases}
0 & \textrm{if} \ k \neq k' \\
\cos^2 \big(\frac{\theta - \theta'}{2}\big) & \textrm{if} \ k = k'
\end{cases}
\end{equation}

This metric assigns a score of 0 to rings that settle to different cycle types, or rings with the same cycle type but are completely out of phase with each other ($|\theta - \theta'| = \pi$). For rings of the same phase, similarity increases as $|\theta - \theta'| \to 0$, with a maximum similarity of 1 only if $\theta = \theta'$.

We use this similarity metric to define the \textit{correlation function} that measures the average similarity between rings as a function of their (Manhattan) distance from each other in the lattice. Letting $\mathcal{N}(d) = \{(i,j), (m,n) | |i-m| + |j-n| = d\}$ denote the set of pairs of lattices sites a distance $d$ apart from each other, the correlation function $C(d)$ is defined to be the average
\begin{equation}
C(d) = \frac{1}{|\mathcal{N}(d)|} \sum_{(i,j),(m,n) \in \mathcal{N}(d)} \langle k_{ij},\theta_{ij}; k_{mn}, \theta_{mn} \rangle
\end{equation}

We numerically estimate $C(d)$ from simulation data and fit it to an exponential decay $C(d) \propto e^{-d/\xi}$. The fit parameter $\xi$ is called the \textit{correlation length}, and it characterizes the rate at which the similarity between rings decreases. Larger $\xi$ means ring similarity decays less quickly with distance, so the average region size is larger as $\xi$ increases.

The process of fitting the exponential model $C(d) \propto e^{-d/\xi}$ to simulation data is performed via least-squares regression. For instance, we take a logarithm to obtain the linear equation $\log C(d) = -d/\xi + \delta$ and extract the correlation length $\xi$ from slope of the least-squares regression line.

\subsection{Dynamics of Homogeneous Lattices}

We find that lattices of rings with even number of neurons evolve towards most of the rings in the network being in a specific cycle type. This \textit{dominant} cycle type $k_{\textrm{dom}}$ depends on the lattice connectivity $L, T, R,$ \& $B$, but $k_{\textrm{dom}}$ is never far from the most saturated cycle type $N/2$. Fig. \ref{fig:phase-patterns} shows example states of 100x100 lattices at time $t \gg 0$ for different lattice configurations.  

\textbf{Small rings} --- those with $N=4, 6$, and some with $N=8$ neurons ---  tend towards the fully saturated cycle type $k = N/2$ (c.f. Fig. \ref{fig:phase-patterns}a-c).  These rings that converge to this dominant periodic orbit form large contiguous regions --- herein referred to as \textit{domains} --- separated by thin boundaries of rings with non-dominant cycle type. Phase plots of the rings with dominant cycle type $N/2$ show that each domain consists of rings with nearly the same phases $\theta_{ij}$ and crossing a domain boundary coincides with rapid phase change. The qualitative structure of these domains depends on the ring size and lattice configuration parameters. Some configurations form rectangular regions, whereas the domains in other lattices tend to be less structured.

The time-dependence of the domains varies by the connectivity template of the network. Based on our initialization these domains are small for small $t$; they grow in size as the simulation progresses; and the steady-state size of the domains depends on lattice configuration. We find that 4-ring lattices settle in finite time to the global periodic orbit with all rings being synchronized in the fully saturated 2-cycle. This result holds even for larger lattices of size 250x250. 6-ring and 8-ring lattices also see the formation of large phase domains, but the size of these domains appears to saturate for 100x100 networks, even at times up to $10^6 \tau$. 

\begin{figure*}[!t]
   \includegraphics[width=1.9\columnwidth]{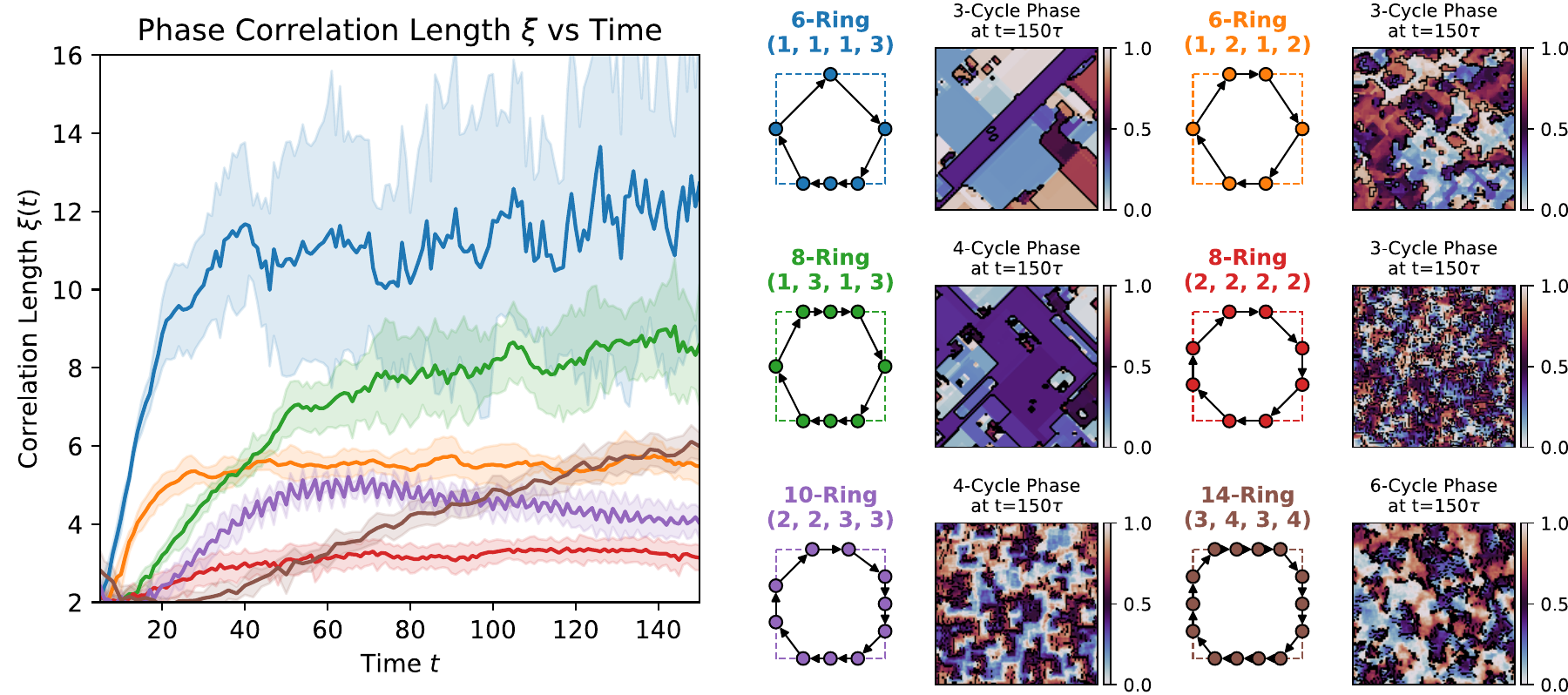}
   \caption{
     \label{fig:small-ring-correlations}
     \textbf{Correlation length dynamics} \ \ 
     (left) Correlation length $\xi$ vs. time for six different lattice connectivity templates $(L, T, R, B)$, with the color-to-connectivity mapping as indicated in the diagrams on the (\textit{right}).
     For each lattice, the solid curve indicates the average correlation length at each point of time, taken over 10 different initial states for the lattice.
     The shaded regions indicate one standard derivation from the means.
     On (\textit{right}), we show both the ring connectivity template $(L, T, R, B)$ and a snapshot of the phases $\theta$ of a specific cycle type at time $t \gg 0$, starting from a random initial state.
     These snapshots give a qualitative sense of scale for the final phase domains of these networks, whose scale is measured \textit{quantitatively} using correlation lengths in the (left) subplot.  
     For the given initialization scheme (30\% neurons firing initially), early correlations appear to be independent of the ring size, with $\xi \approx 2$ for small times in all networks. The growth of the phase domains coincides with $\xi$ increasingly steadily during the first 10-40$\tau$. The correlation length eventually plateaus afterwards, with the steady-state value of $\xi$ being dependent on the connectivity template $(L,T,R,B)$ for the network. 
     The lattices $(L,T,R,B)=(1,1,1,3)$ and $(L,T,R,B)=(1,3,1,3)$ grow to correlation lengths $\xi > 10$, whereas the other lattices form correlated domains of smaller scale. 
     The rate at which the correlation rate saturates seems to depend primarily on the ring size, with larger rates requiring more time to saturate. For instance, the lattice of rings of size 14 does not seem to saturate within the 150$\tau$ time units simulated here.
     These results demonstrate a \textit{spectrum} of phase correlation dynamics available to ring oscillator lattices, ranging from highly correlated to only local synchronization.
    }
\end{figure*}

Computing the time-dependence of the phase correlation length in these networks makes these statements more quantitative. Fig. \ref{fig:small-ring-correlations} plots the correlation length vs. time, averaged across multiple initial states, for some 6-ring and 8-ring oscillator networks. For the given initialization scheme (30\% neurons firing initially), early correlations appears to be independent of the ring size, with $\xi \approx 2$ for small times in all networks. The growth of the phase domains coincides with $\xi$ increasing steadily during the first 10-20$\tau$. The correlation length eventually plateaus afterwards, with the steady-state value of $\xi$ being dependent on the connectivity template $(L,T,R,B)$ for the network.

\textbf{Larger rings} with 10 or more neurons present a wider deviations in dynamics. Most of these networks settle to a dominant cycle and and form (nearly)-equi-phase domains, but the dominant cycle type is no longer $N/2$ and the domains are not necessarily separated by clear domain boundaries of rings in non-dominant cycle types. Many lattice configurations with $N \ge 10$ produce domains resembling labyrinths like those in Fig. \ref{fig:phase-patterns}d.

We found that rings of size 14 with $(L,T,R,B) = (3,4,3,4)$ behavior similarly to the smaller rings, forming large domains of nearly the same phase. The dominant cycle type for these lattices where $k_\textrm{dom} = 6$, and these lattices tend to saturate to a larger correlation length, as suggested in Fig. \ref{fig:phase-patterns}e

For the 16-ring lattice configuration $L=T=R=B=4$, we found the lattice settles to a steady state in which \textit{multiple} cycle types persist simultaneously (Fig. \ref{fig:phase-patterns}e). Indeed, we see large domains formed by rings with cycle types $k = 6$ and $k = 7$, with cycle type $k=6$ being the most prevalent. Both cycle type coexist at least up to time $t = 10^5 \tau$ used for the simulations.

%
%
\section{CONCLUSION}


In this work, we employed numerical simulation to characterize the dynamics of recurrent neural networks built from differentiating neurons. Unlike their integrating counterparts traditionally used in artificial neural networks, differentiating neurons are most responsive to rapid changes in input, and this behavior gives rise to periodic orbits when these neurons are organized into rings. We found that the number of stable periodic orbits grows linearly with the ring size, with one stable orbit per number of allowed pulses stored in the ring. Ring trajectories were characterized by a pulse separation property, where the distance between pulses in the increases over time as the ring converges to one of the stable orbits.  

We also found in simulation that these ring oscillators, when carefully coupled into lattices that we call \textit{homogeneous}, form regions of oscillator synchronization, akin to domains in magnetic spin systems. Both the scale and structure of these phase-correlated regions in the steady-state and the rate at which they develop varied widely with the coupling geometry. Indeed, for certain geometries, rings of small and large sizes are capable of achieving nearly global synchronization, at least in the lattice sizes studied here, and for other geometries, the same ring sizes saturated to only short-range correlations. 


The dependence of oscillator phase synchronization on coupling geometry suggests these networks of differentiating ring oscillator might be tuned to be used as reservoir computers (RCs) \citep{Maass2002RC, Jaeger2001TheechoST}.
This technique employs a fixed but randomly generated dynamical system called a \textit{reservoir} to apply a nonlinear mapping to the input sequences, lifting them to higher dimensions wherein a simple model, often a linear readout, is applied to translate the transformed input to the network's output. 
This framework is general in that any random driven dynamical system might function as the reservoir, including physical systems or networks of coupled oscillators \citep{Kim18, Kim19}, but care must be taken in selecting the appropriate substrate.
Indeed, in avoiding backpropagation through time entirely, the burden is placed on the randomly initialized reservoir to exhibit sufficiently complex responses to input. 
Existing literature does not seem to agree on what makes a reservoir ``sufficiently complex,'' though some experiments suggest optimal reservoirs are those operating at the boundary between ordered and chaotic behavior, often called the ``edge of chaos'' \citep{Legenstein2007EdgeOC, Baranok2014MemoryCO, Dambre2012InformationPC, Langton1990ComputationAT}.

Existing literature \citep{Kim18, Kim19} found that other oscillator networks used as reservoirs performed optimally in parameter regimes at the ``edge of synchronization,'' similar to the notion of \textit{criticality} arising more generally in the RC literature.
Future work with differentiating neurons thus is needed to better understand the effects of geometry on synchronization, ideally in a parameterized way that allows for direct control over the phase-correlation scale in the network. 


This work can also be extended by incorporating both differentiating and integrating neurons into the same networks. 
Biologically brains exhibit even more diversity in neuronal function, 
and such neural heterogeneity has been found to improve learning performance in artificial spiking networks \citep{PerezNieves2020NeuralHP}.
We propose that such mixed integrator-differentiator networks might improve the performance of modern machine learning models even outside of reservoir computing, such as in control problems where differential networks were initially used. 
Further work is needed to find a suitable finite-difference time-stepping method for these neurons that integrate them into gradient-based training frameworks, akin to differential equation based neural network models \citep{Chen2018NeuralOD}.
Using the event-based simulation introduced here in Section \ref{sec:nv-nets-intro}, mixed integrator-differentiator networks might also be designed via heuristic optimization algorithms such as neuro-evolution \citep{Stanley2002EvolvingNN} and its variants, as these approaches have potential to identify effective strategies for coupling different types of neurons together.

%
%
\section{Acknowledgments}

This paper has been many years in the making. We'd like to thank all those who've added their insights to the project, especially Jon Machta, Mark Tilden, AJ Uppal, AJ Yeung, Mohsin Shah, Manor Askenazi, Jack Kenney, Adam Kohan, \& Jason Zheung.

%
%
\section{Author Declarations}

\subsection*{Conflict of Interest}

\noindent The authors have no conflicts of interest to disclose.

\subsection*{Author Contributions}

\noindent \textbf{Peter DelMastro}: Conceptualization (equal), Methodology \& Implementation, Original draft, Review \& editing (equal). \
\textbf{Arjun Karuvally}: Review \& editing (equal). \ 
\textbf{Hananel Hazen}: Review \& editing (equal). \ 
\textbf{Hava Siegelmann}: Review \& editing (equal). \
\textbf{Ed Rietman}: Conceptualization (equal), Supervision, Review \& editing (equal). 

%
%
\section{References}
\bibliography{references}

\begin{thebibliography}{46}%
\makeatletter
\providecommand \@ifxundefined [1]{%
 \@ifx{#1\undefined}
}%
\providecommand \@ifnum [1]{%
 \ifnum #1\expandafter \@firstoftwo
 \else \expandafter \@secondoftwo
 \fi
}%
\providecommand \@ifx [1]{%
 \ifx #1\expandafter \@firstoftwo
 \else \expandafter \@secondoftwo
 \fi
}%
\providecommand \natexlab [1]{#1}%
\providecommand \enquote  [1]{``#1''}%
\providecommand \bibnamefont  [1]{#1}%
\providecommand \bibfnamefont [1]{#1}%
\providecommand \citenamefont [1]{#1}%
\providecommand \href@noop [0]{\@secondoftwo}%
\providecommand \href [0]{\begingroup \@sanitize@url \@href}%
\providecommand \@href[1]{\@@startlink{#1}\@@href}%
\providecommand \@@href[1]{\endgroup#1\@@endlink}%
\providecommand \@sanitize@url [0]{\catcode `\\12\catcode `\$12\catcode `\&12\catcode `\#12\catcode `\^12\catcode `\_12\catcode `\%12\relax}%
\providecommand \@@startlink[1]{}%
\providecommand \@@endlink[0]{}%
\providecommand \url  [0]{\begingroup\@sanitize@url \@url }%
\providecommand \@url [1]{\endgroup\@href {#1}{\urlprefix }}%
\providecommand \urlprefix  [0]{URL }%
\providecommand \Eprint [0]{\href }%
\providecommand \doibase [0]{http://dx.doi.org/}%
\providecommand \selectlanguage [0]{\@gobble}%
\providecommand \bibinfo  [0]{\@secondoftwo}%
\providecommand \bibfield  [0]{\@secondoftwo}%
\providecommand \translation [1]{[#1]}%
\providecommand \BibitemOpen [0]{}%
\providecommand \bibitemStop [0]{}%
\providecommand \bibitemNoStop [0]{.\EOS\space}%
\providecommand \EOS [0]{\spacefactor3000\relax}%
\providecommand \BibitemShut  [1]{\csname bibitem#1\endcsname}%
\let\auto@bib@innerbib\@empty
\bibitem [{\citenamefont {Elman}(1990)}]{elman90}%
  \BibitemOpen
  \bibfield  {author} {\bibinfo {author} {\bibfnamefont {J.~L.}\ \bibnamefont {Elman}},\ }\bibfield  {title} {\enquote {\bibinfo {title} {{Finding Structure in Time}},}\ }\href {\doibase https://doi.org/10.1207/s15516709cog1402\_1} {\bibfield  {journal} {\bibinfo  {journal} {Cognitive Science}\ }\textbf {\bibinfo {volume} {14}},\ \bibinfo {pages} {179--211} (\bibinfo {year} {1990})}\BibitemShut {NoStop}%
\bibitem [{\citenamefont {Hochreiter}\ and\ \citenamefont {Schmidhuber}(1997)}]{hochreiter97}%
  \BibitemOpen
  \bibfield  {author} {\bibinfo {author} {\bibfnamefont {S.}~\bibnamefont {Hochreiter}}\ and\ \bibinfo {author} {\bibfnamefont {J.}~\bibnamefont {Schmidhuber}},\ }\bibfield  {title} {\enquote {\bibinfo {title} {Long short-term memory},}\ }\href@noop {} {\bibfield  {journal} {\bibinfo  {journal} {Neural Computation}\ } (\bibinfo {year} {1997})}\BibitemShut {NoStop}%
\bibitem [{\citenamefont {Erichson}\ \emph {et~al.}(2021)\citenamefont {Erichson}, \citenamefont {Azencot}, \citenamefont {Queiruga}, \citenamefont {Hodgkinson},\ and\ \citenamefont {Mahoney}}]{Erichson2021LipschitzRN}%
  \BibitemOpen
  \bibfield  {author} {\bibinfo {author} {\bibfnamefont {N.~B.}\ \bibnamefont {Erichson}}, \bibinfo {author} {\bibfnamefont {O.}~\bibnamefont {Azencot}}, \bibinfo {author} {\bibfnamefont {A.~F.}\ \bibnamefont {Queiruga}}, \bibinfo {author} {\bibfnamefont {L.}~\bibnamefont {Hodgkinson}}, \ and\ \bibinfo {author} {\bibfnamefont {M.~W.}\ \bibnamefont {Mahoney}},\ }\bibfield  {title} {\enquote {\bibinfo {title} {Lipschitz recurrent neural networks},}\ }in\ \href {https://openreview.net/forum?id=-N7PBXqOUJZ} {\emph {\bibinfo {booktitle} {9th International Conference on Learning Representations, {ICLR} 2021, Virtual Event, Austria, May 3-7, 2021}}}\ (\bibinfo  {publisher} {OpenReview.net},\ \bibinfo {year} {2021})\BibitemShut {NoStop}%
\bibitem [{\citenamefont {Rusch}\ and\ \citenamefont {Mishra}(2021)}]{Rusch2020CoupledOR}%
  \BibitemOpen
  \bibfield  {author} {\bibinfo {author} {\bibfnamefont {T.~K.}\ \bibnamefont {Rusch}}\ and\ \bibinfo {author} {\bibfnamefont {S.}~\bibnamefont {Mishra}},\ }\bibfield  {title} {\enquote {\bibinfo {title} {Coupled oscillatory recurrent neural network (cornn): An accurate and (gradient) stable architecture for learning long time dependencies},}\ }in\ \href {https://openreview.net/forum?id=F3s69XzWOia} {\emph {\bibinfo {booktitle} {9th International Conference on Learning Representations, {ICLR} 2021, Virtual Event, Austria, May 3-7, 2021}}}\ (\bibinfo  {publisher} {OpenReview.net},\ \bibinfo {year} {2021})\BibitemShut {NoStop}%
\bibitem [{\citenamefont {Rusch}\ \emph {et~al.}(2022)\citenamefont {Rusch}, \citenamefont {Mishra}, \citenamefont {Erichson},\ and\ \citenamefont {Mahoney}}]{Rusch2022LongEM}%
  \BibitemOpen
  \bibfield  {author} {\bibinfo {author} {\bibfnamefont {T.~K.}\ \bibnamefont {Rusch}}, \bibinfo {author} {\bibfnamefont {S.}~\bibnamefont {Mishra}}, \bibinfo {author} {\bibfnamefont {N.~B.}\ \bibnamefont {Erichson}}, \ and\ \bibinfo {author} {\bibfnamefont {M.~W.}\ \bibnamefont {Mahoney}},\ }\bibfield  {title} {\enquote {\bibinfo {title} {Long expressive memory for sequence modeling},}\ }in\ \href {https://openreview.net/forum?id=vwj6aUeocyf} {\emph {\bibinfo {booktitle} {The Tenth International Conference on Learning Representations, {ICLR} 2022, Virtual Event, April 25-29, 2022}}}\ (\bibinfo  {publisher} {OpenReview.net},\ \bibinfo {year} {2022})\BibitemShut {NoStop}%
\bibitem [{\citenamefont {Gu}, \citenamefont {Goel},\ and\ \citenamefont {R{\'{e}}}(2022)}]{Gu2022S4}%
  \BibitemOpen
  \bibfield  {author} {\bibinfo {author} {\bibfnamefont {A.}~\bibnamefont {Gu}}, \bibinfo {author} {\bibfnamefont {K.}~\bibnamefont {Goel}}, \ and\ \bibinfo {author} {\bibfnamefont {C.}~\bibnamefont {R{\'{e}}}},\ }\bibfield  {title} {\enquote {\bibinfo {title} {Efficiently modeling long sequences with structured state spaces},}\ }in\ \href {https://openreview.net/forum?id=uYLFoz1vlAC} {\emph {\bibinfo {booktitle} {The Tenth International Conference on Learning Representations, {ICLR} 2022, Virtual Event, April 25-29, 2022}}}\ (\bibinfo  {publisher} {OpenReview.net},\ \bibinfo {year} {2022})\BibitemShut {NoStop}%
\bibitem [{\citenamefont {Chandra}, \citenamefont {Goyal},\ and\ \citenamefont {Gupta}(2021)}]{Chandra2021EvaluationOD}%
  \BibitemOpen
  \bibfield  {author} {\bibinfo {author} {\bibfnamefont {R.}~\bibnamefont {Chandra}}, \bibinfo {author} {\bibfnamefont {S.}~\bibnamefont {Goyal}}, \ and\ \bibinfo {author} {\bibfnamefont {R.}~\bibnamefont {Gupta}},\ }\bibfield  {title} {\enquote {\bibinfo {title} {Evaluation of deep learning models for multi-step ahead time series prediction},}\ }\href@noop {} {\bibfield  {journal} {\bibinfo  {journal} {IEEE Access}\ }\textbf {\bibinfo {volume} {9}},\ \bibinfo {pages} {83105--83123} (\bibinfo {year} {2021})}\BibitemShut {NoStop}%
\bibitem [{\citenamefont {Pathak}\ \emph {et~al.}(2018)\citenamefont {Pathak}, \citenamefont {Hunt}, \citenamefont {Girvan}, \citenamefont {Lu},\ and\ \citenamefont {Ott}}]{Pathak2018ModelFreePO}%
  \BibitemOpen
  \bibfield  {author} {\bibinfo {author} {\bibfnamefont {J.}~\bibnamefont {Pathak}}, \bibinfo {author} {\bibfnamefont {B.~R.}\ \bibnamefont {Hunt}}, \bibinfo {author} {\bibfnamefont {M.}~\bibnamefont {Girvan}}, \bibinfo {author} {\bibfnamefont {Z.}~\bibnamefont {Lu}}, \ and\ \bibinfo {author} {\bibfnamefont {E.}~\bibnamefont {Ott}},\ }\bibfield  {title} {\enquote {\bibinfo {title} {Model-free prediction of large spatiotemporally chaotic systems from data: A reservoir computing approach.}}\ }\href@noop {} {\bibfield  {journal} {\bibinfo  {journal} {Physical review letters}\ }\textbf {\bibinfo {volume} {120 2}},\ \bibinfo {pages} {024102} (\bibinfo {year} {2018})}\BibitemShut {NoStop}%
\bibitem [{\citenamefont {Melis}, \citenamefont {Dyer},\ and\ \citenamefont {Blunsom}(2017)}]{Melis2017OnTS}%
  \BibitemOpen
  \bibfield  {author} {\bibinfo {author} {\bibfnamefont {G.}~\bibnamefont {Melis}}, \bibinfo {author} {\bibfnamefont {C.}~\bibnamefont {Dyer}}, \ and\ \bibinfo {author} {\bibfnamefont {P.}~\bibnamefont {Blunsom}},\ }\bibfield  {title} {\enquote {\bibinfo {title} {On the state of the art of evaluation in neural language models},}\ }\href@noop {} {\bibfield  {journal} {\bibinfo  {journal} {ArXiv}\ }\textbf {\bibinfo {volume} {abs/1707.05589}} (\bibinfo {year} {2017})}\BibitemShut {NoStop}%
\bibitem [{\citenamefont {Merity}, \citenamefont {Keskar},\ and\ \citenamefont {Socher}(2017)}]{Merity2017RegularizingAO}%
  \BibitemOpen
  \bibfield  {author} {\bibinfo {author} {\bibfnamefont {S.}~\bibnamefont {Merity}}, \bibinfo {author} {\bibfnamefont {N.~S.}\ \bibnamefont {Keskar}}, \ and\ \bibinfo {author} {\bibfnamefont {R.}~\bibnamefont {Socher}},\ }\bibfield  {title} {\enquote {\bibinfo {title} {Regularizing and optimizing lstm language models},}\ }\href@noop {} {\bibfield  {journal} {\bibinfo  {journal} {ArXiv}\ }\textbf {\bibinfo {volume} {abs/1708.02182}} (\bibinfo {year} {2017})}\BibitemShut {NoStop}%
\bibitem [{\citenamefont {Gu}\ and\ \citenamefont {Dao}(2023)}]{Gu2023MambaLS}%
  \BibitemOpen
  \bibfield  {author} {\bibinfo {author} {\bibfnamefont {A.}~\bibnamefont {Gu}}\ and\ \bibinfo {author} {\bibfnamefont {T.}~\bibnamefont {Dao}},\ }\bibfield  {title} {\enquote {\bibinfo {title} {Mamba: Linear-time sequence modeling with selective state spaces},}\ }\href@noop {} {\bibfield  {journal} {\bibinfo  {journal} {ArXiv}\ }\textbf {\bibinfo {volume} {abs/2312.00752}} (\bibinfo {year} {2023})}\BibitemShut {NoStop}%
\bibitem [{\citenamefont {Meng}, \citenamefont {Gorbet},\ and\ \citenamefont {Kuli'c}(2021)}]{Meng2021MemorybasedDR}%
  \BibitemOpen
  \bibfield  {author} {\bibinfo {author} {\bibfnamefont {L.}~\bibnamefont {Meng}}, \bibinfo {author} {\bibfnamefont {R.~B.}\ \bibnamefont {Gorbet}}, \ and\ \bibinfo {author} {\bibfnamefont {D.}~\bibnamefont {Kuli'c}},\ }\bibfield  {title} {\enquote {\bibinfo {title} {Memory-based deep reinforcement learning for pomdps},}\ }\href@noop {} {\bibfield  {journal} {\bibinfo  {journal} {2021 IEEE/RSJ International Conference on Intelligent Robots and Systems (IROS)}\ ,\ \bibinfo {pages} {5619--5626}} (\bibinfo {year} {2021})}\BibitemShut {NoStop}%
\bibitem [{\citenamefont {Jaeger}\ \emph {et~al.}(2007)\citenamefont {Jaeger}, \citenamefont {Lukoševičius}, \citenamefont {Popovici},\ and\ \citenamefont {Siewert}}]{Jaeger2007LeakyIntegrator}%
  \BibitemOpen
  \bibfield  {author} {\bibinfo {author} {\bibfnamefont {H.}~\bibnamefont {Jaeger}}, \bibinfo {author} {\bibfnamefont {M.}~\bibnamefont {Lukoševičius}}, \bibinfo {author} {\bibfnamefont {D.}~\bibnamefont {Popovici}}, \ and\ \bibinfo {author} {\bibfnamefont {U.}~\bibnamefont {Siewert}},\ }\bibfield  {title} {\enquote {\bibinfo {title} {2007 special issue: Optimization and applications of echo state networks with leaky- integrator neurons},}\ }\href {https://api.semanticscholar.org/CorpusID:17795421} {\bibfield  {journal} {\bibinfo  {journal} {Neural Networks}\ }\textbf {\bibinfo {volume} {20}},\ \bibinfo {pages} {335--352} (\bibinfo {year} {2007})}\BibitemShut {NoStop}%
\bibitem [{\citenamefont {Hopfield}(1982)}]{Hopfield1982NeuralNA}%
  \BibitemOpen
  \bibfield  {author} {\bibinfo {author} {\bibfnamefont {J.~J.}\ \bibnamefont {Hopfield}},\ }\bibfield  {title} {\enquote {\bibinfo {title} {Neural networks and physical systems with emergent collective computational abilities.}}\ }\href@noop {} {\bibfield  {journal} {\bibinfo  {journal} {Proceedings of the National Academy of Sciences of the United States of America}\ }\textbf {\bibinfo {volume} {79 8}},\ \bibinfo {pages} {2554--8} (\bibinfo {year} {1982})}\BibitemShut {NoStop}%
\bibitem [{\citenamefont {Krotov}\ and\ \citenamefont {Hopfield}(2021)}]{Krotov2021LargeAM}%
  \BibitemOpen
  \bibfield  {author} {\bibinfo {author} {\bibfnamefont {D.}~\bibnamefont {Krotov}}\ and\ \bibinfo {author} {\bibfnamefont {J.~J.}\ \bibnamefont {Hopfield}},\ }\bibfield  {title} {\enquote {\bibinfo {title} {Large associative memory problem in neurobiology and machine learning},}\ }in\ \href {https://openreview.net/forum?id=X4y\_10OX-hX} {\emph {\bibinfo {booktitle} {9th International Conference on Learning Representations, {ICLR} 2021, Virtual Event, Austria, May 3-7, 2021}}}\ (\bibinfo  {publisher} {OpenReview.net},\ \bibinfo {year} {2021})\BibitemShut {NoStop}%
\bibitem [{\citenamefont {Karuvally}, \citenamefont {Sejnowski},\ and\ \citenamefont {Siegelmann}(2022)}]{Karuvally2022EnergybasedGS}%
  \BibitemOpen
  \bibfield  {author} {\bibinfo {author} {\bibfnamefont {A.}~\bibnamefont {Karuvally}}, \bibinfo {author} {\bibfnamefont {T.~J.}\ \bibnamefont {Sejnowski}}, \ and\ \bibinfo {author} {\bibfnamefont {H.~T.}\ \bibnamefont {Siegelmann}},\ }\bibfield  {title} {\enquote {\bibinfo {title} {Energy-based general sequential episodic memory networks at the adiabatic limit},}\ }\href {\doibase 10.48550/ARXIV.2212.05563} {\bibfield  {journal} {\bibinfo  {journal} {CoRR}\ }\textbf {\bibinfo {volume} {abs/2212.05563}} (\bibinfo {year} {2022}),\ 10.48550/ARXIV.2212.05563},\ \Eprint {http://arxiv.org/abs/2212.05563} {2212.05563} \BibitemShut {NoStop}%
\bibitem [{\citenamefont {Doya}(1993)}]{Doya1993}%
  \BibitemOpen
  \bibfield  {author} {\bibinfo {author} {\bibfnamefont {K.}~\bibnamefont {Doya}},\ }\bibfield  {title} {\enquote {\bibinfo {title} {Bifurcations of recurrent neural networks in gradient descent learning},}\ }\href@noop {} {\bibfield  {journal} {\bibinfo  {journal} {IEEE Transactions on Neural Networks}\ } (\bibinfo {year} {1993})}\BibitemShut {NoStop}%
\bibitem [{\citenamefont {Bengio}, \citenamefont {Simard},\ and\ \citenamefont {Frasconi}(1994)}]{bengio94}%
  \BibitemOpen
  \bibfield  {author} {\bibinfo {author} {\bibfnamefont {Y.}~\bibnamefont {Bengio}}, \bibinfo {author} {\bibfnamefont {P.}~\bibnamefont {Simard}}, \ and\ \bibinfo {author} {\bibfnamefont {P.}~\bibnamefont {Frasconi}},\ }\bibfield  {title} {\enquote {\bibinfo {title} {Learning long-term dependencies with gradient descent is difficult},}\ }\href {\doibase 10.1109/72.279181} {\bibfield  {journal} {\bibinfo  {journal} {IEEE transactions on neural networks / a publication of the IEEE Neural Networks Council}\ }\textbf {\bibinfo {volume} {5}},\ \bibinfo {pages} {157--66} (\bibinfo {year} {1994})}\BibitemShut {NoStop}%
\bibitem [{\citenamefont {Pascanu}, \citenamefont {Mikolov},\ and\ \citenamefont {Bengio}(2012)}]{Pascanu2012OnTD}%
  \BibitemOpen
  \bibfield  {author} {\bibinfo {author} {\bibfnamefont {R.}~\bibnamefont {Pascanu}}, \bibinfo {author} {\bibfnamefont {T.}~\bibnamefont {Mikolov}}, \ and\ \bibinfo {author} {\bibfnamefont {Y.}~\bibnamefont {Bengio}},\ }\bibfield  {title} {\enquote {\bibinfo {title} {On the difficulty of training recurrent neural networks},}\ }in\ \href {https://api.semanticscholar.org/CorpusID:14650762} {\emph {\bibinfo {booktitle} {International Conference on Machine Learning}}}\ (\bibinfo {year} {2012})\BibitemShut {NoStop}%
\bibitem [{\citenamefont {Stiefel}\ and\ \citenamefont {Coggan}(2023)}]{Stiefel2023TheEC}%
  \BibitemOpen
  \bibfield  {author} {\bibinfo {author} {\bibfnamefont {K.~M.}\ \bibnamefont {Stiefel}}\ and\ \bibinfo {author} {\bibfnamefont {J.~S.}\ \bibnamefont {Coggan}},\ }\bibfield  {title} {\enquote {\bibinfo {title} {The energy challenges of artificial superintelligence},}\ }\href {https://api.semanticscholar.org/CorpusID:264497615} {\bibfield  {journal} {\bibinfo  {journal} {Frontiers in Artificial Intelligence}\ }\textbf {\bibinfo {volume} {6}} (\bibinfo {year} {2023})}\BibitemShut {NoStop}%
\bibitem [{\citenamefont {Hasslacher}\ and\ \citenamefont {Tilden}(1997)}]{Hasslacher97}%
  \BibitemOpen
  \bibfield  {author} {\bibinfo {author} {\bibfnamefont {B.}~\bibnamefont {Hasslacher}}\ and\ \bibinfo {author} {\bibfnamefont {M.~W.}\ \bibnamefont {Tilden}},\ }\bibfield  {title} {\enquote {\bibinfo {title} {{Theoretical foundations for nervous networks}},}\ }\href {\doibase 10.1063/1.54209} {\bibfield  {journal} {\bibinfo  {journal} {AIP Conference Proceedings}\ }\textbf {\bibinfo {volume} {411}},\ \bibinfo {pages} {179--184} (\bibinfo {year} {1997})}\BibitemShut {NoStop}%
\bibitem [{\citenamefont {Rietman}, \citenamefont {Tilden},\ and\ \citenamefont {Askenazi}(2003)}]{Rietman03}%
  \BibitemOpen
  \bibfield  {author} {\bibinfo {author} {\bibfnamefont {E.~A.}\ \bibnamefont {Rietman}}, \bibinfo {author} {\bibfnamefont {M.~W.}\ \bibnamefont {Tilden}}, \ and\ \bibinfo {author} {\bibfnamefont {M.}~\bibnamefont {Askenazi}},\ }\bibfield  {title} {\enquote {\bibinfo {title} {Analog computation with rings of quasiperiodic oscillators: the microdynamics of cognition in living machines},}\ }\href@noop {} {\bibfield  {journal} {\bibinfo  {journal} {Robotics Auton. Syst.}\ }\textbf {\bibinfo {volume} {45}},\ \bibinfo {pages} {249--263} (\bibinfo {year} {2003})}\BibitemShut {NoStop}%
\bibitem [{\citenamefont {Frigo}\ and\ \citenamefont {Tilden}(1995)}]{SATBOT1995}%
  \BibitemOpen
  \bibfield  {author} {\bibinfo {author} {\bibfnamefont {J.}~\bibnamefont {Frigo}}\ and\ \bibinfo {author} {\bibfnamefont {M.~W.}\ \bibnamefont {Tilden}},\ }\bibfield  {title} {\enquote {\bibinfo {title} {Satbot i: Prototype of a biomorphic autonomous spacecraft},}\ }\href {https://www.osti.gov/biblio/150940} {\  (\bibinfo {year} {1995})}\BibitemShut {NoStop}%
\bibitem [{\citenamefont {Frigo}\ and\ \citenamefont {Tilden}(1998)}]{Frigo98}%
  \BibitemOpen
  \bibfield  {author} {\bibinfo {author} {\bibfnamefont {J.~R.}\ \bibnamefont {Frigo}}\ and\ \bibinfo {author} {\bibfnamefont {M.~W.}\ \bibnamefont {Tilden}},\ }\bibfield  {title} {\enquote {\bibinfo {title} {Biologically inspired neural network controller for an infrared tracking system},}\ }in\ \href {\doibase 10.1117/12.335726} {\emph {\bibinfo {booktitle} {Mobile Robots {XIII} and Intelligent Transportation Systems, Boston, MA, USA, November 1, 1998}}},\ \bibinfo {series} {{SPIE} Proceedings}, Vol.\ \bibinfo {volume} {3525},\ \bibinfo {editor} {edited by\ \bibinfo {editor} {\bibfnamefont {H.}~\bibnamefont {Choset}}, \bibinfo {editor} {\bibfnamefont {D.~W.}\ \bibnamefont {Gage}}, \bibinfo {editor} {\bibfnamefont {P.}~\bibnamefont {Kachroo}}, \bibinfo {editor} {\bibfnamefont {M.}~\bibnamefont {Kourjanski}}, \ and\ \bibinfo {editor} {\bibfnamefont {M.~J.}\ \bibnamefont {de~Vries}}}\ (\bibinfo  {publisher} {{SPIE}},\ \bibinfo {year} {1998})\ pp.\ \bibinfo {pages} {393--403}\BibitemShut {NoStop}%
\bibitem [{\citenamefont {Hasslacher}\ and\ \citenamefont {Tilden}(1995)}]{Hasslacher95}%
  \BibitemOpen
  \bibfield  {author} {\bibinfo {author} {\bibfnamefont {B.}~\bibnamefont {Hasslacher}}\ and\ \bibinfo {author} {\bibfnamefont {M.~W.}\ \bibnamefont {Tilden}},\ }\bibfield  {title} {\enquote {\bibinfo {title} {Living machines},}\ }\href {\doibase 10.1016/0921-8890(95)00019-C} {\bibfield  {journal} {\bibinfo  {journal} {Robotics Auton. Syst.}\ }\textbf {\bibinfo {volume} {15}},\ \bibinfo {pages} {143--169} (\bibinfo {year} {1995})}\BibitemShut {NoStop}%
\bibitem [{\citenamefont {Still}\ and\ \citenamefont {Tilden}(1998)}]{Still98}%
  \BibitemOpen
  \bibfield  {author} {\bibinfo {author} {\bibfnamefont {S.}~\bibnamefont {Still}}\ and\ \bibinfo {author} {\bibfnamefont {M.~W.}\ \bibnamefont {Tilden}},\ }\enquote {\bibinfo {title} {Controller for a four-legged walking machine},}\ in\ \href {\doibase 10.1142/9789812816535\_0012} {\emph {\bibinfo {booktitle} {Neuromorphic Systems}}}\ (\bibinfo {year} {1998})\ pp.\ \bibinfo {pages} {138--148},\ \Eprint {http://arxiv.org/abs/https://www.worldscientific.com/doi/pdf/10.1142/9789812816535\_0012} {https://www.worldscientific.com/doi/pdf/10.1142/9789812816535\_0012} \BibitemShut {NoStop}%
\bibitem [{\citenamefont {Rodan}\ and\ \citenamefont {Tiňo}(2011)}]{Rodan2011MinimumComplexESN}%
  \BibitemOpen
  \bibfield  {author} {\bibinfo {author} {\bibfnamefont {A.}~\bibnamefont {Rodan}}\ and\ \bibinfo {author} {\bibfnamefont {P.}~\bibnamefont {Tiňo}},\ }\bibfield  {title} {\enquote {\bibinfo {title} {Minimum complexity echo state network},}\ }\href {https://api.semanticscholar.org/CorpusID:319654} {\bibfield  {journal} {\bibinfo  {journal} {IEEE Transactions on Neural Networks}\ }\textbf {\bibinfo {volume} {22}},\ \bibinfo {pages} {131--144} (\bibinfo {year} {2011})}\BibitemShut {NoStop}%
\bibitem [{\citenamefont {Sethna}(2021)}]{Sethna2021StatMechCorrFunc}%
  \BibitemOpen
  \bibfield  {author} {\bibinfo {author} {\bibfnamefont {J.~P.}\ \bibnamefont {Sethna}},\ }\enquote {\bibinfo {title} {Statistical mechanics: Entropy, order parameters, and complexity},}\ \ (\bibinfo  {publisher} {Oxford Press},\ \bibinfo {year} {2021})\ Chap.\ \bibinfo {chapter} {10: Correlations, response, and dissipation},\ \bibinfo {edition} {2nd}\ ed.\BibitemShut {Stop}%
\bibitem [{\citenamefont {Nakao}(2016)}]{Nakao2016PhaseReduction}%
  \BibitemOpen
  \bibfield  {author} {\bibinfo {author} {\bibfnamefont {H.}~\bibnamefont {Nakao}},\ }\bibfield  {title} {\enquote {\bibinfo {title} {Phase reduction approach to synchronisation of nonlinear oscillators},}\ }\href@noop {} {\bibfield  {journal} {\bibinfo  {journal} {Contemporary Physics}\ }\textbf {\bibinfo {volume} {57}},\ \bibinfo {pages} {188 -- 214} (\bibinfo {year} {2016})}\BibitemShut {NoStop}%
\bibitem [{\citenamefont {Kuramoto}(1975)}]{Kuramoto1975Original}%
  \BibitemOpen
  \bibfield  {author} {\bibinfo {author} {\bibfnamefont {Y.}~\bibnamefont {Kuramoto}},\ }\bibfield  {title} {\enquote {\bibinfo {title} {Self-entrainment of a population of coupled non-linear oscillators},}\ }in\ \href@noop {} {\emph {\bibinfo {booktitle} {International Symposium on Mathematical Problems in Theoretical Physics}}},\ \bibinfo {editor} {edited by\ \bibinfo {editor} {\bibfnamefont {H.}~\bibnamefont {Araki}}}\ (\bibinfo  {publisher} {Springer Berlin Heidelberg},\ \bibinfo {address} {Berlin, Heidelberg},\ \bibinfo {year} {1975})\ pp.\ \bibinfo {pages} {420--422}\BibitemShut {NoStop}%
\bibitem [{\citenamefont {Kuramoto}(1984)}]{Kuramoto1984ChemicalOW}%
  \BibitemOpen
  \bibfield  {author} {\bibinfo {author} {\bibfnamefont {Y.}~\bibnamefont {Kuramoto}},\ }\bibfield  {title} {\enquote {\bibinfo {title} {Chemical oscillations, waves, and turbulence},}\ }in\ \href@noop {} {\emph {\bibinfo {booktitle} {Springer Series in Synergetics}}}\ (\bibinfo {year} {1984})\BibitemShut {NoStop}%
\bibitem [{\citenamefont {Acebr{\'o}n}\ \emph {et~al.}(2005)\citenamefont {Acebr{\'o}n}, \citenamefont {Bonilla}, \citenamefont {Vicente}, \citenamefont {Ritort},\ and\ \citenamefont {Spigler}}]{Acebrn2005KuramotoRev1}%
  \BibitemOpen
  \bibfield  {author} {\bibinfo {author} {\bibfnamefont {J.~A.}\ \bibnamefont {Acebr{\'o}n}}, \bibinfo {author} {\bibfnamefont {L.~L.}\ \bibnamefont {Bonilla}}, \bibinfo {author} {\bibfnamefont {C.~J.~P.}\ \bibnamefont {Vicente}}, \bibinfo {author} {\bibfnamefont {F.}~\bibnamefont {Ritort}}, \ and\ \bibinfo {author} {\bibfnamefont {R.}~\bibnamefont {Spigler}},\ }\bibfield  {title} {\enquote {\bibinfo {title} {The kuramoto model: A simple paradigm for synchronization phenomena},}\ }\href@noop {} {\bibfield  {journal} {\bibinfo  {journal} {Reviews of Modern Physics}\ }\textbf {\bibinfo {volume} {77}},\ \bibinfo {pages} {137--185} (\bibinfo {year} {2005})}\BibitemShut {NoStop}%
\bibitem [{\citenamefont {Rodrigues}\ \emph {et~al.}(2016)\citenamefont {Rodrigues}, \citenamefont {Peron}, \citenamefont {Ji},\ and\ \citenamefont {Kurths}}]{Rodrigues2015KuramotoRev2}%
  \BibitemOpen
  \bibfield  {author} {\bibinfo {author} {\bibfnamefont {F.~A.}\ \bibnamefont {Rodrigues}}, \bibinfo {author} {\bibfnamefont {T.~K.~D.}\ \bibnamefont {Peron}}, \bibinfo {author} {\bibfnamefont {P.}~\bibnamefont {Ji}}, \ and\ \bibinfo {author} {\bibfnamefont {J.}~\bibnamefont {Kurths}},\ }\bibfield  {title} {\enquote {\bibinfo {title} {The kuramoto model in complex networks},}\ }\href {\doibase https://doi.org/10.1016/j.physrep.2015.10.008} {\bibfield  {journal} {\bibinfo  {journal} {Physics Reports}\ }\textbf {\bibinfo {volume} {610}},\ \bibinfo {pages} {1--98} (\bibinfo {year} {2016})},\ \bibinfo {note} {the Kuramoto model in complex networks}\BibitemShut {NoStop}%
\bibitem [{\citenamefont {Sarkar}\ and\ \citenamefont {Gupte}(2021)}]{Sarkar2021KuramotoLattice}%
  \BibitemOpen
  \bibfield  {author} {\bibinfo {author} {\bibfnamefont {M.}~\bibnamefont {Sarkar}}\ and\ \bibinfo {author} {\bibfnamefont {N.}~\bibnamefont {Gupte}},\ }\bibfield  {title} {\enquote {\bibinfo {title} {Phase synchronization in the two-dimensional kuramoto model: Vortices and duality.}}\ }\href@noop {} {\bibfield  {journal} {\bibinfo  {journal} {Physical review. E}\ }\textbf {\bibinfo {volume} {103 3-1}},\ \bibinfo {pages} {032204} (\bibinfo {year} {2021})}\BibitemShut {NoStop}%
\bibitem [{\citenamefont {Sarkar}(2021)}]{Sarkar2021KuramotoLatticeWithNoise}%
  \BibitemOpen
  \bibfield  {author} {\bibinfo {author} {\bibfnamefont {M.}~\bibnamefont {Sarkar}},\ }\bibfield  {title} {\enquote {\bibinfo {title} {Synchronization transition in the two-dimensional kuramoto model with dichotomous noise.}}\ }\href {https://api.semanticscholar.org/CorpusID:236155269} {\bibfield  {journal} {\bibinfo  {journal} {Chaos}\ }\textbf {\bibinfo {volume} {31 8}},\ \bibinfo {pages} {083102} (\bibinfo {year} {2021})}\BibitemShut {NoStop}%
\bibitem [{\citenamefont {Maass}, \citenamefont {Natschl{\"a}ger},\ and\ \citenamefont {Markram}(2002)}]{Maass2002RC}%
  \BibitemOpen
  \bibfield  {author} {\bibinfo {author} {\bibfnamefont {W.}~\bibnamefont {Maass}}, \bibinfo {author} {\bibfnamefont {T.}~\bibnamefont {Natschl{\"a}ger}}, \ and\ \bibinfo {author} {\bibfnamefont {H.}~\bibnamefont {Markram}},\ }\bibfield  {title} {\enquote {\bibinfo {title} {Real-time computing without stable states: A new framework for neural computation based on perturbations},}\ }\href {https://api.semanticscholar.org/CorpusID:1045112} {\bibfield  {journal} {\bibinfo  {journal} {Neural Computation}\ }\textbf {\bibinfo {volume} {14}},\ \bibinfo {pages} {2531--2560} (\bibinfo {year} {2002})}\BibitemShut {NoStop}%
\bibitem [{\citenamefont {Jaeger}(2001)}]{Jaeger2001TheechoST}%
  \BibitemOpen
  \bibfield  {author} {\bibinfo {author} {\bibfnamefont {H.}~\bibnamefont {Jaeger}},\ }\bibfield  {title} {\enquote {\bibinfo {title} {The''echo state''approach to analysing and training recurrent neural networks},}\ \ }(\bibinfo {year} {2001})\BibitemShut {NoStop}%
\bibitem [{\citenamefont {Choi}\ and\ \citenamefont {Kim}(2018)}]{Kim18}%
  \BibitemOpen
  \bibfield  {author} {\bibinfo {author} {\bibfnamefont {J.}~\bibnamefont {Choi}}\ and\ \bibinfo {author} {\bibfnamefont {P.}~\bibnamefont {Kim}},\ }\bibfield  {title} {\enquote {\bibinfo {title} {Critical neuromorphic computing based on explosive synchronization},}\ }\href {http://arxiv.org/abs/1810.10944} {\bibfield  {journal} {\bibinfo  {journal} {CoRR}\ }\textbf {\bibinfo {volume} {abs/1810.10944}} (\bibinfo {year} {2018})},\ \Eprint {http://arxiv.org/abs/1810.10944} {1810.10944} \BibitemShut {NoStop}%
\bibitem [{\citenamefont {Choi}\ and\ \citenamefont {Kim}(2019)}]{Kim19}%
  \BibitemOpen
  \bibfield  {author} {\bibinfo {author} {\bibfnamefont {J.}~\bibnamefont {Choi}}\ and\ \bibinfo {author} {\bibfnamefont {P.}~\bibnamefont {Kim}},\ }\bibfield  {title} {\enquote {\bibinfo {title} {Reservoir computing based on quenched chaos},}\ }\href {http://arxiv.org/abs/1909.01571} {\bibfield  {journal} {\bibinfo  {journal} {CoRR}\ }\textbf {\bibinfo {volume} {abs/1909.01571}} (\bibinfo {year} {2019})},\ \Eprint {http://arxiv.org/abs/1909.01571} {1909.01571} \BibitemShut {NoStop}%
\bibitem [{\citenamefont {Legenstein}\ and\ \citenamefont {Maass}(2007)}]{Legenstein2007EdgeOC}%
  \BibitemOpen
  \bibfield  {author} {\bibinfo {author} {\bibfnamefont {R.~A.}\ \bibnamefont {Legenstein}}\ and\ \bibinfo {author} {\bibfnamefont {W.}~\bibnamefont {Maass}},\ }\bibfield  {title} {\enquote {\bibinfo {title} {Edge of chaos and prediction of computational performance for neural circuit models},}\ }\href {https://api.semanticscholar.org/CorpusID:7249094} {\bibfield  {journal} {\bibinfo  {journal} {Neural networks : the official journal of the International Neural Network Society}\ }\textbf {\bibinfo {volume} {20 3}},\ \bibinfo {pages} {323--34} (\bibinfo {year} {2007})}\BibitemShut {NoStop}%
\bibitem [{\citenamefont {Baranvok}\ and\ \citenamefont {Farkas}(2014)}]{Baranok2014MemoryCO}%
  \BibitemOpen
  \bibfield  {author} {\bibinfo {author} {\bibfnamefont {P.}~\bibnamefont {Baranvok}}\ and\ \bibinfo {author} {\bibfnamefont {I.}~\bibnamefont {Farkas}},\ }\bibfield  {title} {\enquote {\bibinfo {title} {Memory capacity of input-driven echo state networks at the edge of chaos},}\ }in\ \href {https://api.semanticscholar.org/CorpusID:12748419} {\emph {\bibinfo {booktitle} {International Conference on Artificial Neural Networks}}}\ (\bibinfo {year} {2014})\BibitemShut {NoStop}%
\bibitem [{\citenamefont {Dambre}\ \emph {et~al.}(2012)\citenamefont {Dambre}, \citenamefont {Verstraeten}, \citenamefont {Schrauwen},\ and\ \citenamefont {Massar}}]{Dambre2012InformationPC}%
  \BibitemOpen
  \bibfield  {author} {\bibinfo {author} {\bibfnamefont {J.}~\bibnamefont {Dambre}}, \bibinfo {author} {\bibfnamefont {D.}~\bibnamefont {Verstraeten}}, \bibinfo {author} {\bibfnamefont {B.}~\bibnamefont {Schrauwen}}, \ and\ \bibinfo {author} {\bibfnamefont {S.}~\bibnamefont {Massar}},\ }\bibfield  {title} {\enquote {\bibinfo {title} {Information processing capacity of dynamical systems},}\ }\href {https://api.semanticscholar.org/CorpusID:7342429} {\bibfield  {journal} {\bibinfo  {journal} {Scientific Reports}\ }\textbf {\bibinfo {volume} {2}} (\bibinfo {year} {2012})}\BibitemShut {NoStop}%
\bibitem [{\citenamefont {Langton}(1990)}]{Langton1990ComputationAT}%
  \BibitemOpen
  \bibfield  {author} {\bibinfo {author} {\bibfnamefont {C.~G.}\ \bibnamefont {Langton}},\ }\bibfield  {title} {\enquote {\bibinfo {title} {Computation at the edge of chaos: Phase transitions and emergent computation},}\ }\href {https://api.semanticscholar.org/CorpusID:62374310} {\bibfield  {journal} {\bibinfo  {journal} {Physica D: Nonlinear Phenomena}\ }\textbf {\bibinfo {volume} {42}},\ \bibinfo {pages} {12--37} (\bibinfo {year} {1990})}\BibitemShut {NoStop}%
\bibitem [{\citenamefont {Perez-Nieves}\ \emph {et~al.}(2020)\citenamefont {Perez-Nieves}, \citenamefont {Leung}, \citenamefont {Dragotti},\ and\ \citenamefont {Goodman}}]{PerezNieves2020NeuralHP}%
  \BibitemOpen
  \bibfield  {author} {\bibinfo {author} {\bibfnamefont {N.}~\bibnamefont {Perez-Nieves}}, \bibinfo {author} {\bibfnamefont {V.~C.~H.}\ \bibnamefont {Leung}}, \bibinfo {author} {\bibfnamefont {P.~L.}\ \bibnamefont {Dragotti}}, \ and\ \bibinfo {author} {\bibfnamefont {D.~F.~M.}\ \bibnamefont {Goodman}},\ }\bibfield  {title} {\enquote {\bibinfo {title} {Neural heterogeneity promotes robust learning},}\ }\href@noop {} {\bibfield  {journal} {\bibinfo  {journal} {Nature Communications}\ }\textbf {\bibinfo {volume} {12}} (\bibinfo {year} {2020})}\BibitemShut {NoStop}%
\bibitem [{\citenamefont {Chen}\ \emph {et~al.}(2018)\citenamefont {Chen}, \citenamefont {Rubanova}, \citenamefont {Bettencourt},\ and\ \citenamefont {Duvenaud}}]{Chen2018NeuralOD}%
  \BibitemOpen
  \bibfield  {author} {\bibinfo {author} {\bibfnamefont {T.~Q.}\ \bibnamefont {Chen}}, \bibinfo {author} {\bibfnamefont {Y.}~\bibnamefont {Rubanova}}, \bibinfo {author} {\bibfnamefont {J.}~\bibnamefont {Bettencourt}}, \ and\ \bibinfo {author} {\bibfnamefont {D.~K.}\ \bibnamefont {Duvenaud}},\ }\bibfield  {title} {\enquote {\bibinfo {title} {Neural ordinary differential equations},}\ }in\ \href@noop {} {\emph {\bibinfo {booktitle} {Neural Information Processing Systems}}}\ (\bibinfo {year} {2018})\BibitemShut {NoStop}%
\bibitem [{\citenamefont {Stanley}\ and\ \citenamefont {Miikkulainen}(2002)}]{Stanley2002EvolvingNN}%
  \BibitemOpen
  \bibfield  {author} {\bibinfo {author} {\bibfnamefont {K.~O.}\ \bibnamefont {Stanley}}\ and\ \bibinfo {author} {\bibfnamefont {R.}~\bibnamefont {Miikkulainen}},\ }\bibfield  {title} {\enquote {\bibinfo {title} {Evolving neural networks through augmenting topologies},}\ }\href@noop {} {\bibfield  {journal} {\bibinfo  {journal} {Evolutionary Computation}\ }\textbf {\bibinfo {volume} {10}},\ \bibinfo {pages} {99--127} (\bibinfo {year} {2002})}\BibitemShut {NoStop}%
\end{thebibliography}%

\end{document}